\documentclass[runningheads,a4paper]{llncs}

\usepackage{amssymb}
\usepackage{graphicx}
\usepackage{xcolor}
\usepackage{url}
\usepackage{tikz}
\usetikzlibrary{arrows,decorations.pathmorphing,backgrounds,fit,positioning,shapes.symbols,chains}
\usepackage{amsmath}
\usepackage{booktabs} 
\usepackage{multirow}
\usepackage{subfig}
\usepackage{listings}

\urldef{\mailsa}\path|{bh1511, bkainz, a.alansary14, steven.mcdonagh, | 
\urldef{\mailsb}\path|d.rueckert, b.glocker }@imperial.ac.uk|

\graphicspath{{./}}

\newcommand{\todo}[1]{}
\renewcommand{\todo}[1]{{\color{red} TODO: {#1}}}

\begin{document}

\mainmatter  

\title{Predicting Slice-to-Volume Transformation in Presence of Arbitrary Subject Motion}

\titlerunning{Predicting Slice-to-Volume Transformation}

%
%
\author{Benjamin Hou$^1$, Amir Alansary$^1$, Steven McDonagh$^1$, Alice Davidson$^2$, Mary Rutherford$^2$, Jo V. Hajnal$^2$, Daniel Rueckert$^1$, Ben Glocker$^1$, \\ and Bernhard Kainz$^{1,2}$}
\authorrunning{B. Hou et al.}

\institute{
	$^1$Biomedical Image Analysis Group, Imperial College London \\
	$^2$Division of Imaging Sciences and Biomedical Engineering, King’s College London
}

%

\maketitle

\begin{abstract}

This paper aims to solve a fundamental problem in intensity-based 2D/3D registration, which concerns the limited capture range and need for very good initialization of state-of-the-art image registration methods. We propose a regression approach that learns to predict rotation and translations of arbitrary 2D image slices from 3D volumes, with respect to a learned canonical atlas co-ordinate system. To this end, we utilize Convolutional Neural Networks (CNNs) to learn the highly complex regression function that maps 2D image slices into their correct position and orientation in 3D space. Our approach is attractive in challenging imaging scenarios, where significant subject motion complicates reconstruction performance of 3D volumes from 2D slice data. We extensively evaluate the effectiveness of our approach quantitatively on simulated MRI brain data with extreme random motion. 
We further demonstrate qualitative results on fetal MRI where our method is integrated into a full reconstruction and motion compensation pipeline. With our CNN regression approach we obtain an average prediction error of 7mm on simulated data, and convincing reconstruction quality of images of very young fetuses where previous methods fail. We further discuss applications to Computed Tomography and X-ray projections. Our approach is a general solution to the 2D/3D initialization problem. It is computationally efficient, with prediction times per slice of a few milliseconds, making it suitable for real-time scenarios.

\end{abstract}

\section{Introduction} 

\parskip0pt
Intensity-based registration requires a good initial alignment. General optimisation methods often cannot find a global minimum from any given starting position on the cost function. Thus, image analysis that requires registration, \emph{e.g.}, atlas-based segmentation~\cite{Aljabar2009726}, motion-compensation~\cite{rousseau2006registration}, tracking~\cite{Miao2016}, or clinical analysis of the data visualised in a standard co-ordinate system, often requires manual initalisation of the alignment. This problem gets particularity challenging for applications where the alignment is not defined by a 3D-3D rigid-body transformation. An initial rigid registration can be achieved by selecting common landmarks~\cite{Ghesu2016}. However, many applications, in particular motion compensation techniques, require at least approximate spatial alignment and 3D consistency between individual 2D slices to provide a useful initialisation for subsequent automatic registration methods. Manual alignment of hundreds of slices is not feasible in practice. Landmark-based techniques can mitigate this problem, but is heavily dependent on detection accuracy and robustness of the calculated homography between locations and the descriptive power of the used landmark encoding. 2D slices also do not provide the required 3D information to establish robust landmark matching, therefore this technique cannot be used on applications such as motion compensation in fetal imaging.   

Robustness of (semi-)automatic registration methods is characterised by their \emph{capture range}, which is the maximum transformation offset from which a specific method can recover good spatial alignment. For all currently known intensity-based registration methods, the \emph{capture range} is limited. 

\noindent\textbf{Contribution:} 
We introduce a method that automatically learns slice transformation parameters relative to a canonical atlas co-ordinate system, purely from the encoded intensity information in 2D slices. We propose a Convolution Neural Network (CNN) regression approach that is able to predict and re-orient arbitrarily sampled slices, to provide an accurate initialisation for subsequent intensity-based registration. 
Our method is applicable to a number of clinical situations. In particular, we quantitatively evaluate the prediction performance with simulated 2D slice data extracted from adult 3D MRI brain and thorax phantoms. In addition, we qualitatively evaluate the approach for a full reconstruction and motion compensation pipeline for fetal MRI.
Our approach can naturally be generalised to 3D/3D volumetric registration by predicting the transformation of a few selected slices. It is also applicable to projective images, which is highly valuable for X-ray/CT registration.



\noindent\textbf{Related Work:} 
Slice-to-Volume registration is a key step in medical imaging, as it allows single or multiple 2D images to be registered together in a common world co-ordinate system to form a consistent 3D volume. This provides better visualisation for the practitioner to either diagnose or perform operative procedures. Furthermore, it paves the way to exploit 3D medical image analysis techniques.

In literature one can distinguish between volume-to-slice and slice-to-volume techniques. The first is concerned with aligning a volume to a given image, \emph{e.g.}, aligning an intra-operative C-arm X-ray image to a pre-operative volumetric scan. This can be manually or artificially initialised and many approaches have been proposed to solve this problem. The most advanced solution to this problem we are aware of uses CNNs to evaluate the spatial arrangement of landmarks automatically~\cite{Miao2016}. Besides this, methods that can compensate for large offsets usually require the use of fiducial markers~\cite{Kainz2008}, which makes use of either special equipment or invasive procedures. 

While our method is also applicable to the volume-to-slice problem, as shown in Exp. 3, here we focus on the slice-to-volume problem. Manual alignment of hundreds of slices to each other is much more challenging than the theoretically possible manual initialisation of volume-to-slice problems.  

One target application we discuss in this paper is fetal MRI, where maternal breathing and spontaneous movement from the fetus is a major problem, that involves slice-wise re-alignment of randomly displaced anatomy~\cite{gholipour2010robust,kainz2015fast,rousseau2006registration,kim2010intersection}. Existing methods require good initial spatial consistency between the acquired slices to generate an approximation of the target structure. This approximation is used for iterative refinement of slice-to-volume registration. Good initial 3D slice alignment is only possible trough fast acquisition like single-shot Fast Spin Echo (ssFSE) and the acquisition of temporally close, intersecting stacks of slices. Redundant data covering an area of interest cannot be used from all acquired images since the displacement worsens during the course of an examination, thus redundancy has to be high and, generally, several attempts are necessary to acquire better quality data that can be motion compensated.
Nevertheless, from the clinical practice, we know that individual 2D slices are well examinable and trained experts are able to virtually realign a collection of slices mentally with respect to their real anatomical localization during diagnostics. The recent advent of deep neural network architectures~\cite{lecun2015deep} suggests that such a learning based expert-intuition of slice transformations can also be achieved fully automatically using machine learning.

\section{Method}

\begin{figure}
\vspace{-0.8cm}
\centering
\includegraphics[width=1.0\linewidth]{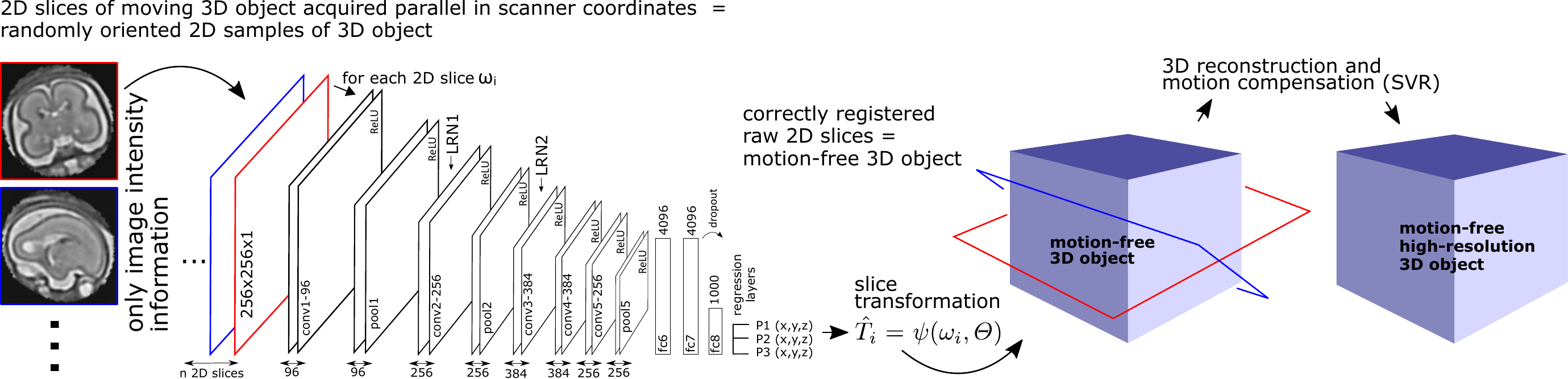}
\caption{Overview over our approach.}
\label{fig:overview}
\vspace{-0.8cm}
\end{figure}

\noindent The core of our method utilises a CNN, called \emph{SVRNet}, to regress and predict transformation parameters $\hat{T_i}$, such that $\hat{T_i} = \psi(\omega_i,\Theta)$, where $\Theta$ is the learned network parameters and $\omega_i \in \Omega$ are a series of 2D image slices
 that are acquired from a moving 3D object $\Omega$.
SVRNet provides a robust initialisation for intensity-based registration refinement by predicting $\hat{T_i}$ for each $\omega_i$ (see Fig.~\ref{fig:overview}). We also define $T_i$ as known ground truth parameters of $\omega_i$ during validation.

Our proposed pipeline consists of three modular components: \textbf{(I)} approximate organ localisation, \textbf{(II)} prediction of $\hat{T_i}$, and \textbf{(III)} 3D reconstruction/ intensity-based registration refinement. 

Organ localisation, which defines a Region of Interest (ROI), can be achieved using rough manual delineation, organ focused scan sequences or automatic methods, such as~\cite{Keraudren2015} for example for the fetal MRI use case. 
For 3D Reconstruction, we use a modified Slice-to-Volume Reconstruction (SVR) method~\cite{kainz2015fast} and initialise it with transformed $\omega_i$ using $\hat{T_i}$. 
Here on, we focus on the novel part of this pipeline, which is SVRNet.
SVRNet needs to be trained accurately on a desired ROI, imaging modality, and use-case scenario. 



\noindent\textbf{Data Set Generation:} $\omega_i$, for training and validation, are generated from $n$ motion free 
3D volumes $\Omega_{train}$. Each volume encloses a desired ROI, is centred at the origin and re-sampled to a cubic volume of length $L$, with spacing $1mm\times1mm\times1mm$. L/4 sampling planes, with spacing of 4mm and size $L \times L$, are evenly spaced along the Z-axis. 

$\omega_i$ at extremities of $\Omega_{train}$
may contain little or no content. If the variance of a particular $\omega_i$ is below a threshold of $t$, where $t = K \cdot \max (\sigma^{2}(\omega_i))$, $\forall i \in \Omega$ and $\sigma^{2}(x) = 1/N\sum_{i}^{N-1}{|x_i - \bar{x}|^{2}}$, then it is omitted. A higher K value will restrict $\omega_i$ to the middle portion of the volume. In our experiments, $K \approx 0.2$, which samples the central 80\% of the volume.


To capture a dense permutation of $\hat{T_i} \in \Omega_{train}$, we rotate the sampling planes about the origin whilst keeping the volume static. Ideally, all rotational permutations should be random and evenly spaced on the surface of a unit sphere. Uniform sampling of polar co-ordinates, $P(\phi,\theta)$, causes denser sampling near the poles. This can lead to an imbalance of training samples. Thus we use Fibonacci sphere sampling~\cite{González2009}, which allows each point to represent approximately the same area. Thus sampling normals can be calculated by $P(\phi_i,\cos^{-1}(z_i))$, where $\phi_i = 2{\pi}i/\Phi$ and $z_i = 1 - (2i+1)/n$, $i \in {0,1,2,...,n-1}$. $\Phi$ is the golden ratio, as $\Phi^{-1}=\Phi-1$, and is defined as $\Phi = (\sqrt{5}+1)/2$.


For both, training and validation, only one hemisphere needs to be sampled due to  symmetry constraints. Sampling planes with normals in the one hemisphere result in the same image as sampling planes with normals in the other hemisphere albeit mirrored.





\noindent\textbf{Ground Truth Labels:} 
$\hat{T_i}$ can be represented by Euler angles (six parameters: $\{r_x,r_y,r_z,t_x,t_y,t_z\}$) or Quaternions (seven parameters: $\{q_1,q_2,q_3,q_4,t_x,t_y,t_z\}$), or by defining  three Cartesian landmarks within the plane (nine parameters). Huynh et al.~\cite{Huynh:2009:MRC:1574521.1574531} have presented detailed analysis on distance functions for 3D rotations. As they are differentiable, we have implemented them as custom loss layers for regressing on rotational parameters. The loss for Euler angles can be expressed as; ${\Psi}_{1}( ({\alpha}_{1},{\beta}_{1},{\gamma}_{1}) , ({\alpha}_{2},{\beta}_{2},{\gamma}_{2})) = \sqrt{ {d({\alpha}_{1},{\alpha}_{2})}^{2} + {d({\beta}_{1},{\beta}_{2})}^{2} + {d({\gamma}_{1},{\gamma}_{2})}^{2} }$ where $ d(a,b) =  \min \{ |a-b| , 2\pi - |a-b| \}$, and $\alpha,\gamma \in [-\pi,\pi); \beta \in [-\pi/2,\pi/2)$. For 
quaternions; ${\Psi}_{2}({q}_{1},{q}_{2}) = \min { \{\|{q}_{1}-{q}_{2}\|,\|{q}_{1}+{q}_{2}\|\} }$, where $q_1$ and $q_2$ are unit quaternions. We have evaluated all of these options and found that the Cartesian landmark approach yielded the highest accuracy. Hence, we use this approach in all our experiments. 
The landmarks can be arbitrarily selected, as long as their location remains consistent for all $\omega_i$. For our experiments, we have chosen the centres of $\omega_i$, $p_c$, and two corners $p_l, p_r$; where $p_c$ = (0,0,z), $p_l$ = $p_c$ + (-L/2,-L/2,0) and $p_r$ = $p_c$ + (L/2,-L/2,0). To take rotation into account, each point is further multiplied by a rotation matrix $R$ to obtain their final position in world co-ordinates.
Each $\omega_i$ can thus be described by nine parameters: $p_c(x,y,z)$, $p_l(x,y,z)$ and $p_r(x,y,z)$.
This approach keeps the nature of the network loss consistent as it only needs to regress in Cartesian co-ordinate space instead of a mixture of Cartesian co-ordinates and rotation parameters. 


\noindent\textbf{Network Design:} 
SVRNet is derived from the CaffeNet~\cite{jia2014caffe} architecture. Experimentation with other architectures has revealed that this approach yields a maximum training performance whilst keeping the training effort feasible.   
For regression, we define multiple loss outputs; one for each $p_c, p_l, p_r$. SVRNet employs therefore a multi-loss framework, which avoids over-fitting to one particular single loss~\cite{Xu:2016:MRD:3024223.3024275}. Fig.~\ref{fig:overview} shows the details of the SVRNet architecture.


\noindent\textbf{3D Reconstruction:} The network predicts $\hat{T_i}$ to certain degree of accuracy. To reconstruct an accurate high-resolution, motion free 3D volume for $\Omega$ from the regression, we integrate an iterative intensity-based SVR motion compensation approach. 
Conventional SVR methods, \emph{e.g.}~\cite{kainz2015fast}, require a certain degree of correct initial 2D slice alignment in scanner co-ordinate space to estimate an initial approximation of a common volume $\Omega$. The approximation of $\Omega$ is subsequently used as a 3D registration target for 2D/3D slice-to-volume registration. Our approach does not depend on good initial slice alignment and disregards slice scanner co-ordinates completely. We only use slice intensity information for SVRNet and generate an initialization for $\Omega$ using the predicted $\hat{T_i}$. We use regularized Super-Resolution and a Point-Spread-Function similar to~\cite{kainz2015fast} to account for  different resolutions of low-resolution ${\omega_i}$ and high-resolution $\Omega$. 
${\omega_i}$-to-$\Omega$ registration is then individually refined using cross-correlation as cost-function and gradient decent for optimization. Optimization uses three scales of a Gaussian Pyramid representation for ${\omega_i}$ and $\Omega$. Robust statistics~\cite{kainz2015fast} identifies ${\omega_i}$ that have been mis-predicted and excludes them from further iterations. 



\vspace{-0.4cm}
\section{Experiments and Results}
\vspace{-0.2cm}
We have tested our approach on 85 randomly selected and accurately segmented healthy adult brains, on a real-world use case scenario with 34 roughly delineated fetal brain MRI scans and on 60 low-dose thorax CT scans with no organ specific segmentation. SVRNet's average prediction error for these datasets is respectively 5.6$\pm$1.07mm, 7.7$\pm$4.80mm, and 5.9$\pm$2.43mm.  
We evaluate 3D reconstruction performance using the Peak Signal-to-Noise Ratio (PSNR) and $\hat{T_i}$ prediction error as average distance in mm between ground truth locations $p_{c,gt}, p_{l,gt}, p_{r,gt}$ and predicted locations $p_{c,p}, p_{l,p}, p_{r,p}$, \emph{i.e.}, $(|| p_{c,gt} - p_{c,p} || + || p_{l,gt} - p_{l,p} || + || p_{r,gt} - p_{r,p} ||)/3.0$.


All experiments are conducted using the Caffe neural network library, on a computer equipped with an Intel 6700K CPU and Nvidia Titan X Pascal GPU. 

\begin{figure}
	\vspace{-0.5cm}
	\centering
	\subfloat[][Original]{\includegraphics[height=2.3cm]{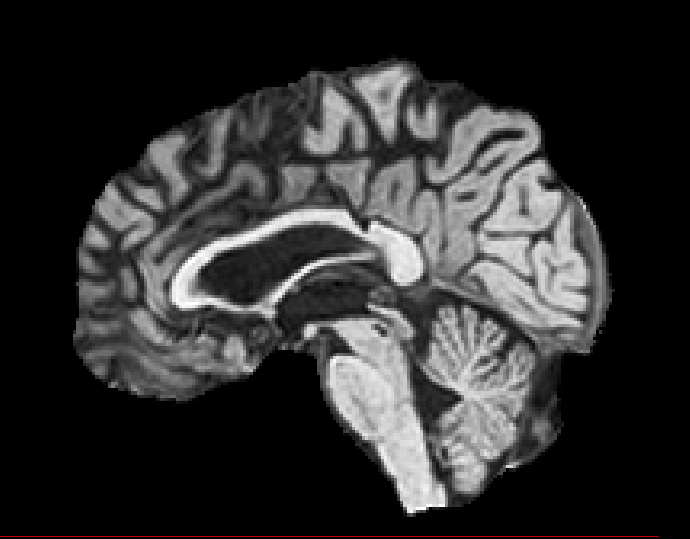}\label{}}
	\subfloat[][SVRNet]{\includegraphics[height=2.3cm]{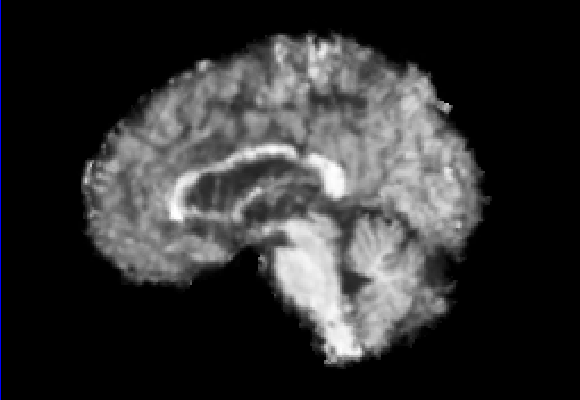}\label{}}
	\subfloat[][+SVR]{\includegraphics[height=2.3cm]{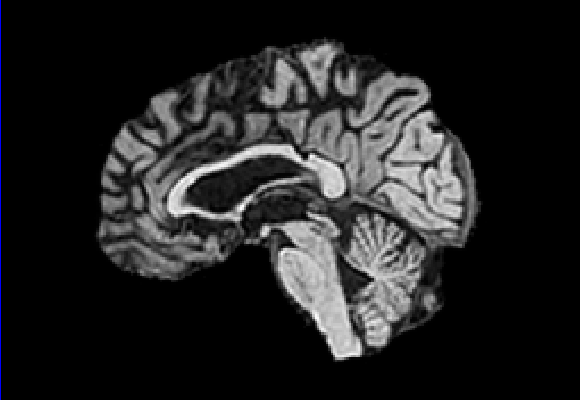}\label{}} 
		\subfloat[][PSNR]{\includegraphics[height=2.3cm]{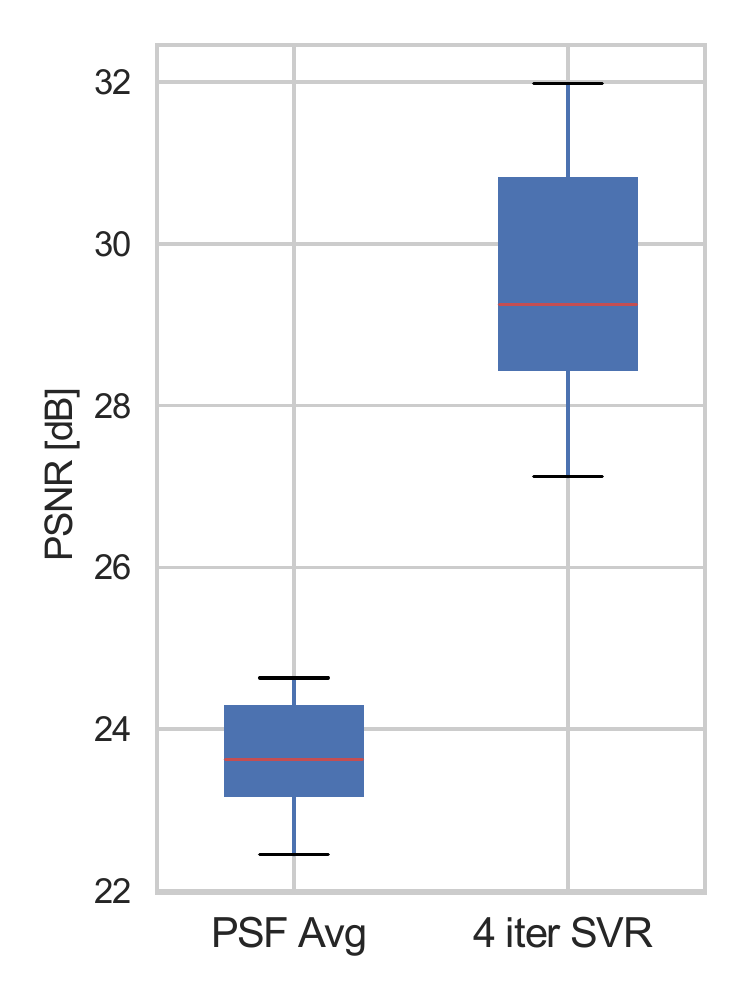}\label{}}
	\caption{Example slice from the segmented adult brain MRI data set (a); reconstruction from 300 $\omega_i$ based on SVRNet regression without SVR (b); SVR initialised with SVRNet predictions after eight iterations of SVR (c). Note that SVRNet (b) predicts individual slice transformations only from image intensities without any initial world co-ordinates of the sampled slice. (d) shows the achieved PSNR in dB when comparing the volumes of (b) and (c) to (a).}
	\label{fig:reconCompADNI}
	\vspace{-0.7cm}
\end{figure}

\noindent\textbf{Exp. 1: Segmented adult brain data} is used to evaluate our network's regression performance with known ground truth $T_i$. 85 brains from the ADNI data set\cite{2_adni} were randomly selected; 70 brains for $\Omega_{train}$ and 15 brains for $\Omega_{validation}$. Fig.~\ref{fig:reconCompADNI} shows an example slice of the ground truth and the reconstructed $\Omega$. 

Each brain has been centered and re-sampled in a $256\times256\times256$ volume. Using the Fibonacci Sphere Sampling method, a density of 500 unique normals is chosen with 64 sampling planes spaced evenly apart on the Z-axis (giving a spacing of 4mm). This therefore yields a maximum of 32000 images per brain; 2.24M for the entire training set and 345K for the entire validation set. After pruning $\omega_i$ with little or no content, this figure drops to approximately 1.2M images for training and 254K for validation. Training took approximately 27hrs for 30 epochs. 

Reconstructing from $\hat{T_i}$ initialisation without SVR yields a PSNR of 23.7 $\pm$ 1.09; with subsequent SVR the PSNR increases to 29.5$\pm$2.43 when tested on 15 randomly selected test volumes after four iterations of SVR.

\begin{figure}[h]
	\centering
	\vspace{-0.7cm}
	\subfloat[][ssFSE ax.]{\includegraphics[height=1.9cm]{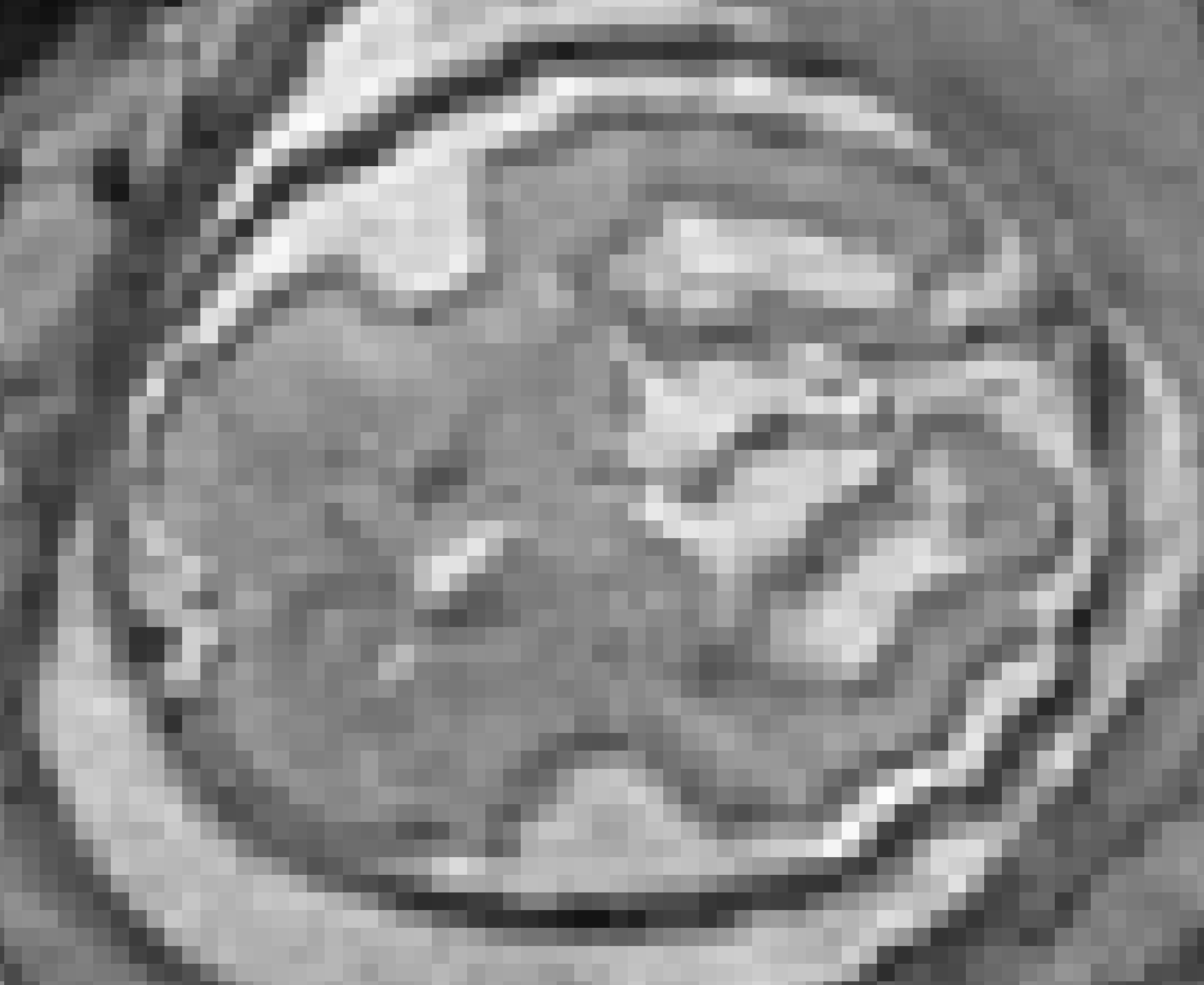}\label{}}
	\subfloat[][ssFSE sag.]{\includegraphics[height=1.9cm]{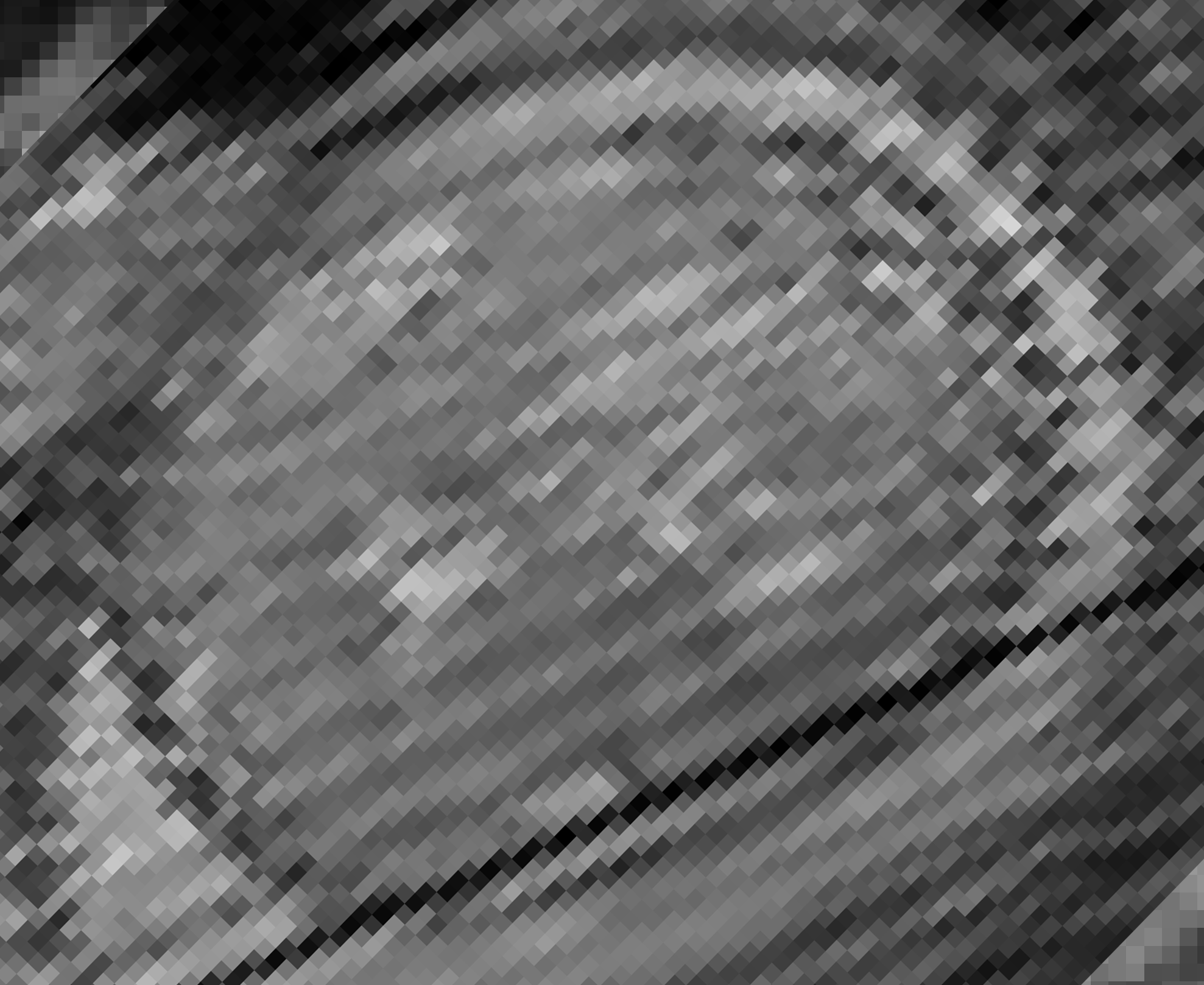}\label{}}
	\subfloat[][train]{\includegraphics[height=1.9cm]{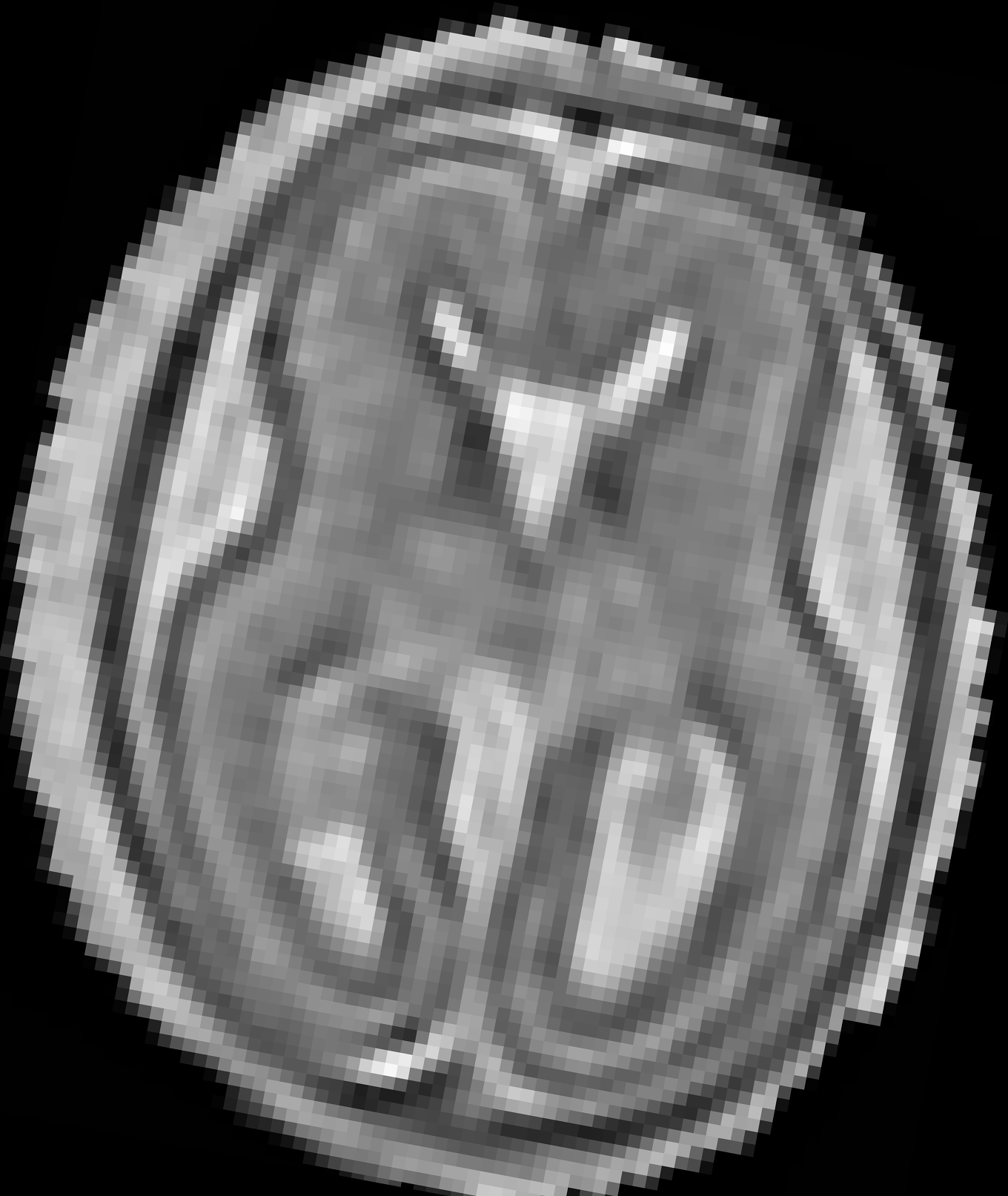}\label{}}
	\subfloat[][SVR]{\includegraphics[height=1.9cm]{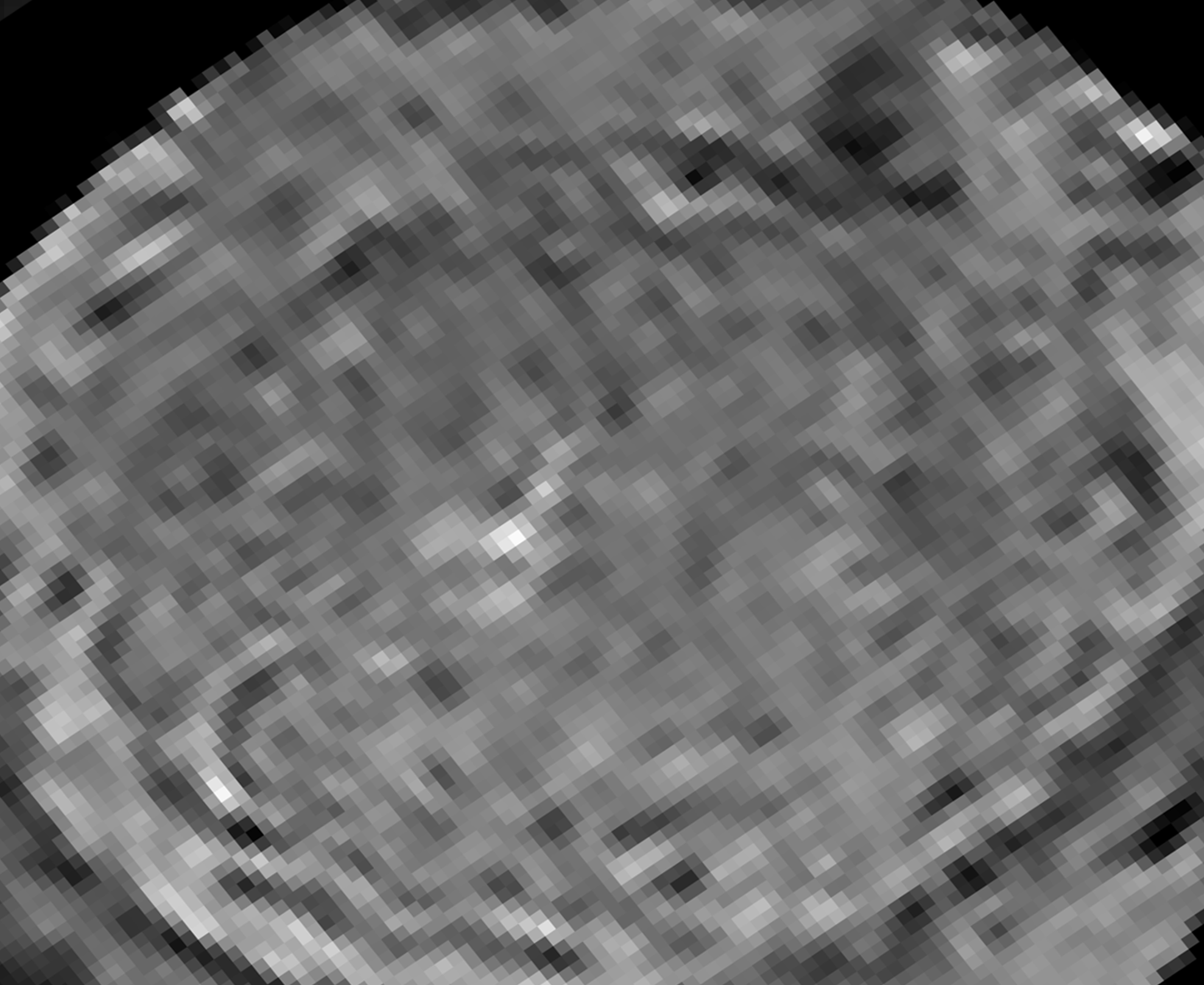}\label{}}
	\subfloat[][SVRNet]{\includegraphics[height=1.9cm]{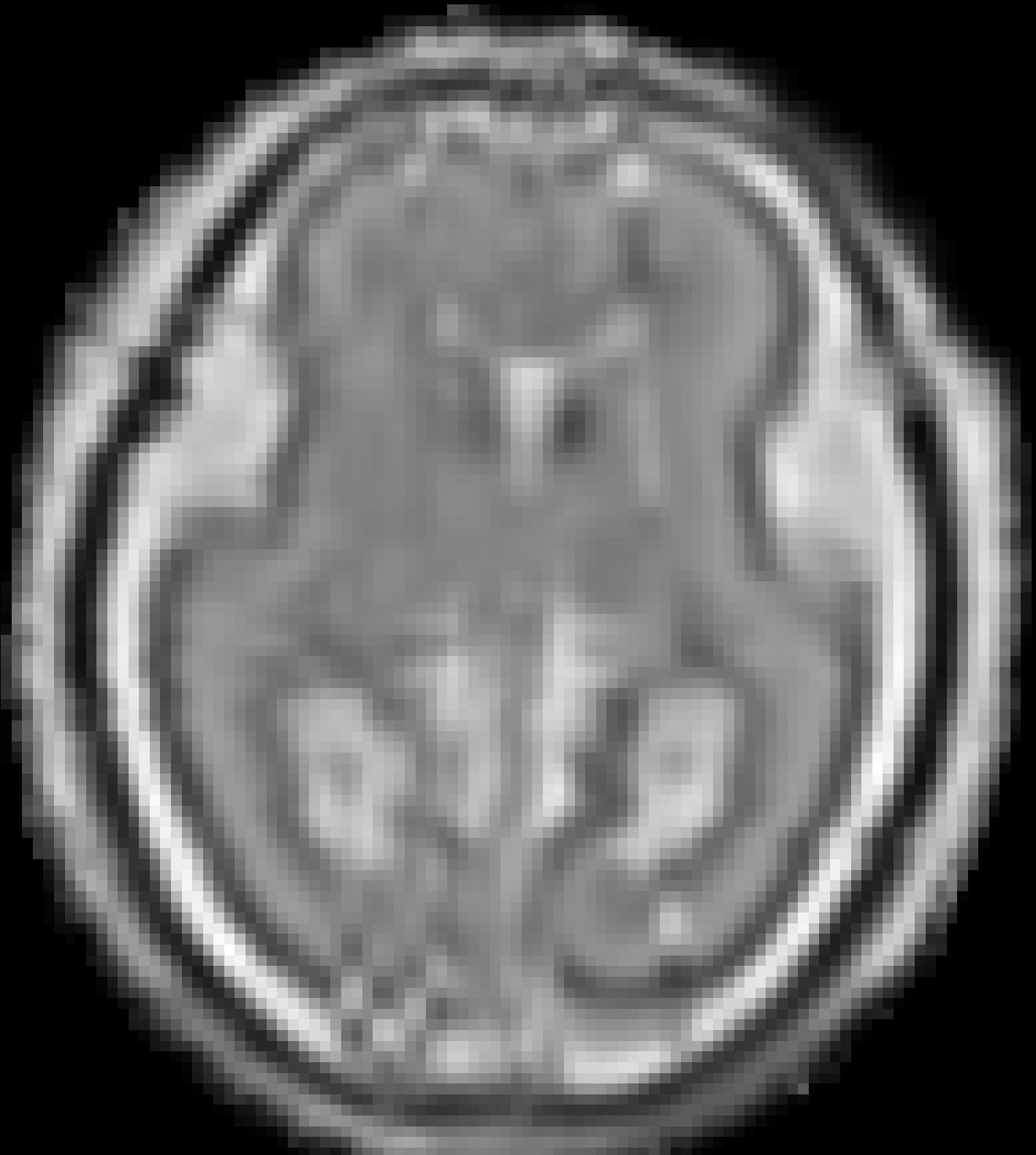}\label{}}
	\subfloat[][+SVR]{\includegraphics[height=1.9cm]{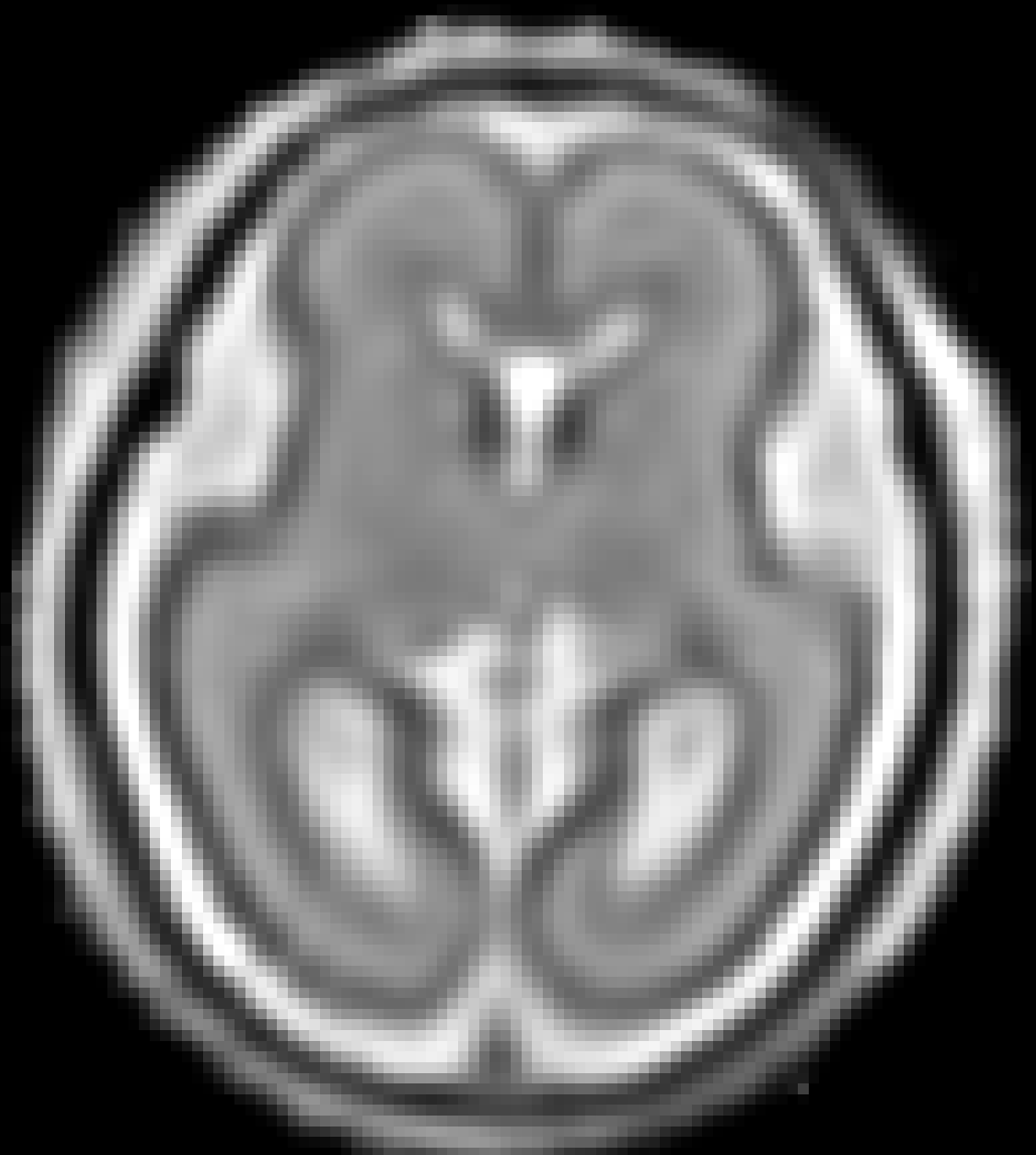}\label{}}
	\caption{Comparison of a single slice from a heavily motion corrupted stack of ssFSE T2 weighted fetal brain MRI (a); axial multi planar reconstruction of one sagittal input stack (b); a slice at approximately the same position through a randomly selected training volume (c); failed reconstruction attempt using standard SVR based on three orthogonal stacks of 2D slices (d) (the fetus moved heavily between the acquisition of the individual stacks); reconstruction based on SVRNet $\hat{T_i}$ regression (e); SVR initialised with SVRNet transformations after eight iterations of SVR (f). Note that (e) and (f) are reconstructed directly in canonical atlas co-ordinates. 
    }
	\label{fig:reconCompFetal}
\end{figure}

\noindent\textbf{Exp. 2: Fetal brain data} is used to test the robustness of our approach under real conditions. Fetuses younger than 30 weeks very often move a lot during examination. Fast MRI sequences allow artifact free acquisition of individual slices but motion between slices corrupts consistent 3D information. 
Fig.~\ref{fig:reconCompFetal} shows that our method is able to accurately predict $\hat{T_i}$ also under these conditions. For this experiment we use ${\omega_i}$ from three orthogonally overlapping stacks of ssFSE slices covering the fetal brain with approximately 20-30 slices each. We are ignoring the stack transformations relative to the scanner and treat each ${\omega_i}$ individually.
For $\Omega_{train}$, 28 clinically approved motion compensated brain reconstructions are resampled into a $150\times150\times150$ volume with $1mm\times1mm\times1mm$ spacing. A density of 500 unique sampling normals has been chosen via the Fibonacci sphere sampling method with 25 sampling planes evenly spaced between -25 to +25 on the Z-axis. This gives a plane spacing of 2mm, sampling only the middle portion of the fetal brain. Training took approximately 10hrs for 30 epochs.
Prediction, \emph{i.e.}, the forward pass through the network, takes approx. 12 ms/slice.

\begin{figure}[h]
	\centering
	\vspace{-0.7cm}
	\subfloat[][original]{\includegraphics[height=1.9cm]{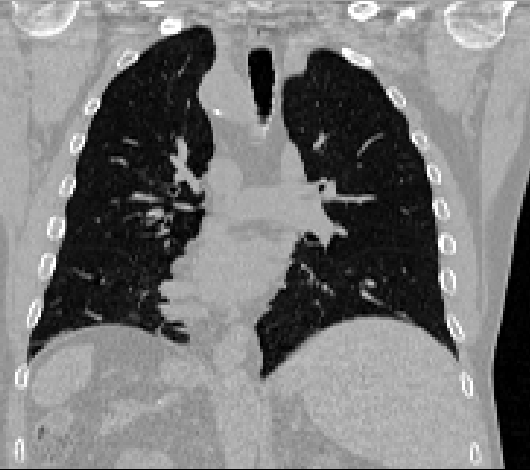}\label{}} 
	\subfloat[][SVRNet]{\includegraphics[height=1.9cm]{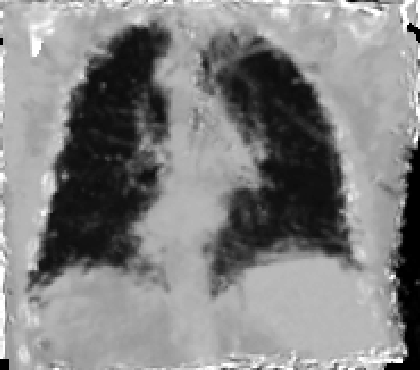}\label{}} 
	\subfloat[][+SVR]{\includegraphics[height=1.9cm]{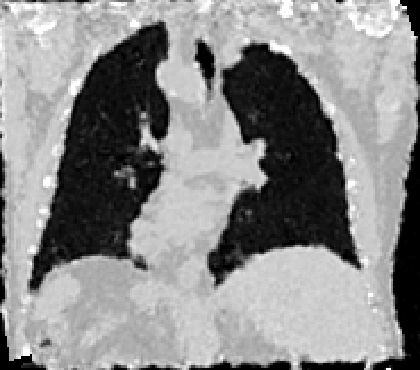}\label{fig:reconCompThoraxc}} 
	\subfloat[][PSNR]{\includegraphics[height=1.9cm]{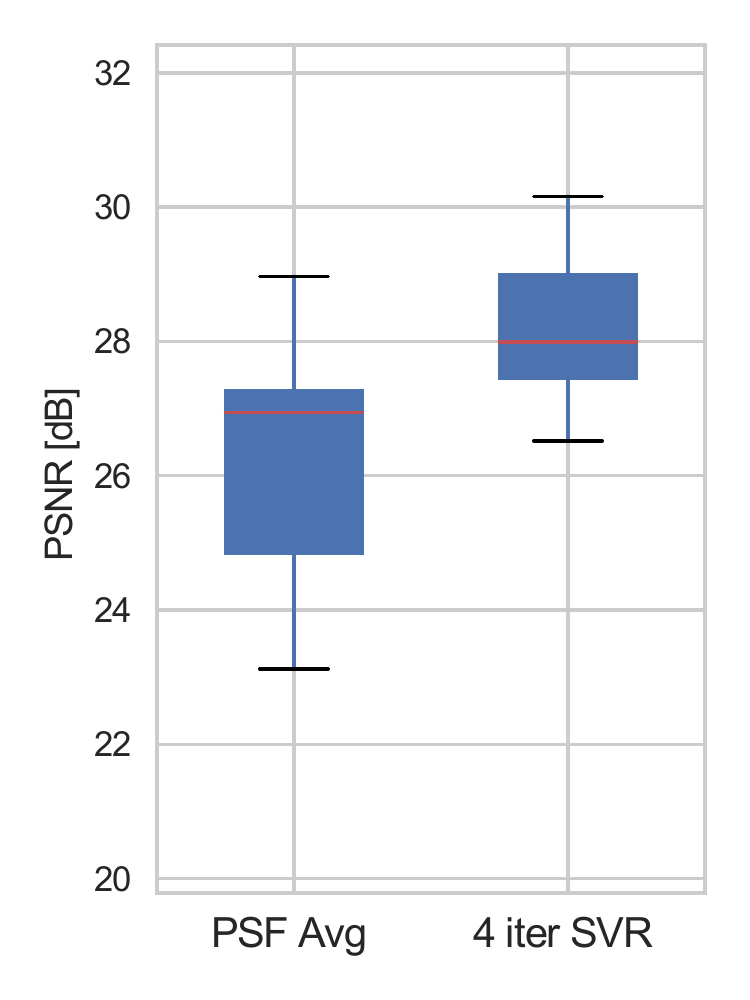}\label{fig:reconCompThoraxcPSNR}} \hfill
	\subfloat[][DRR GT]{\includegraphics[height=1.9cm]{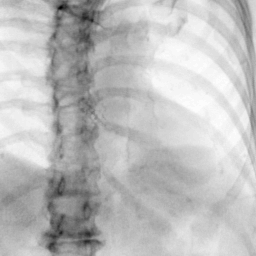}\label{}}  \hfill
	\subfloat[][SVRNet]{\includegraphics[height=1.9cm]{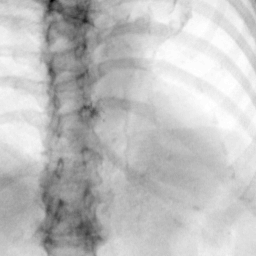}\label{}} \hfill
	\caption{(a): Comparison of a single slice from raw low-dose thorax CT data; (b): reconstruction based on SVRNet $\hat{T_i}$ regression; (c): SVR initialized with SVRNet transformations after four iterations of SVR; (d): PSNR of (b) and (c) compared to (a). (e): shows a projection of an unseen pathological test CT volume as DRR and (e) shows a DRR at the location predicted by our method when presented with the image data in (e).}
	\label{fig:reconCompThorax}
	\vspace{-0.7cm}
\end{figure}

\noindent\textbf{Exp. 3: Adult thorax data:} To show the versatility of our approach we also apply it to adult thorax scans. For this experiment \emph{no organ specific} training is performed but the whole volume is used. We evaluate reconstruction performance similar to Exp. 1 and $\hat{T_i}$ prediction performance when $\Omega$ is projected on an external plane, comparable to X-Ray examination using C-Arms. The latter provides insights about our method's performance when applied to interventional settings in contrast to motion compensation problems.
60 healthy adult thorax scans were randomly selected, 51 scans used for $\Omega_{train}$ and nine scans used for $\Omega_{validation}$. Each scan is intensity normalised and resampled in a volume of $200\times200\times200$ with spacing $1mm\times1mm\times1mm$. Using the Fibonacci sampling method, 25 sampling plane of size $200\times200$, evenly spaced between -50 and +50, were rotated over 500 normals. 
Training took approximately 20 hours for 60 epochs. Fig.~\ref{fig:reconCompThoraxc} shows an example reconstruction result gaining 28dB PSNR with additional SVR. $\hat{T_i}$ prediction takes approx. 20 ms/slice for this data. 

\noindent We use Siddon-Jacobs ray tracing~\cite{Wu2010} to generate Digitally Reconstructed Radiographs (DRRs) from the above described data. For training, we equally sample DRRs on equidistant half-spheres around 51 CT volumes at distances of 80cm, 60cm, and 40cm, between $-90^\circ$ and $90^\circ$ around all three co-ordinate axes. For validation, we generate 1000 DRRs with random rotation parameters within the bounds of the training data at 60cm distance from the volumetric iso-centre. We trained on healthy volunteer data and tested on nine healthy and ten randomly selected pathological volumes (eight lung cancer and two spinal pathologies). 
Our approach is able to predict DRR transformations relative to the trained reference co-ordinate system with an average translation error of 106mm and $5.6^\circ$ plane rotation for healthy patients, and 130mm and $7.0^\circ$ average error for pathological patients. An example is shown in Fig.~\ref{fig:reconCompThorax}e,f. Note that these values are good enough to robustly initialize intensity-based registration refinement. SVRNet prediction can be improved by generating a denser training data set, in particular, in more equidistant half-spheres. 





\noindent\textbf{Discussion \& Conclusion:} 
We have presented a method that is able to predict slice transformations relative to a canonical atlas co-ordinate system. This allows motion compensation for highly motion corrupted scans, \emph{e.g.}, MRI scans of very young fetuses. It allows to incorporate all images that have been acquired during examination and temporal proximity is not required for good initialisation of intensity-based registration methods as it is the case in state-of-the-art methods. We have shown that our method performs remarkably well for fetal brain data in presence of surrounding tissue and without organ specific training for low-dose thorax CT data and X-Ray to CT registration. 

One limitation of our method is that SVRNet requires images to be formatted in the same way the network is trained on. This includes identical intensity ranges, spacing and translation offset removal 
and can be achieved with simple pre-processing methods.
 Furthermore, SVRNet has to be trained for a specific region of interest or organ and scenario (\emph{e.g.}, MRI T1, T2, X-Ray exposure, etc.). 
However, we show that the training region does not need to be delineated accurately and that our method is not restricted with respect to the used imaging modality and scenario. 

\textbf{Acknowledgements:} NVIDIA, Wellcome Trust/EPSRC iFIND [102431], EPSRC EP/N024494/1

\bibliographystyle{splncs03}
\bibliography{refs}


\appendix

\section*{Appendices}

The prediction performance per slice is shown in absolute numbers in Fig.~\ref{fig:adniHistograms}, Fig.~\ref{fig:ThoraxHistograms} and Fig.~\ref{fig:DRRHistograms}.

\subsection{Data generation illustration} 

\begin{figure}[h]
	\centering
	\begin{minipage}[b]{0.3\textwidth}
		\includegraphics[width=\textwidth]{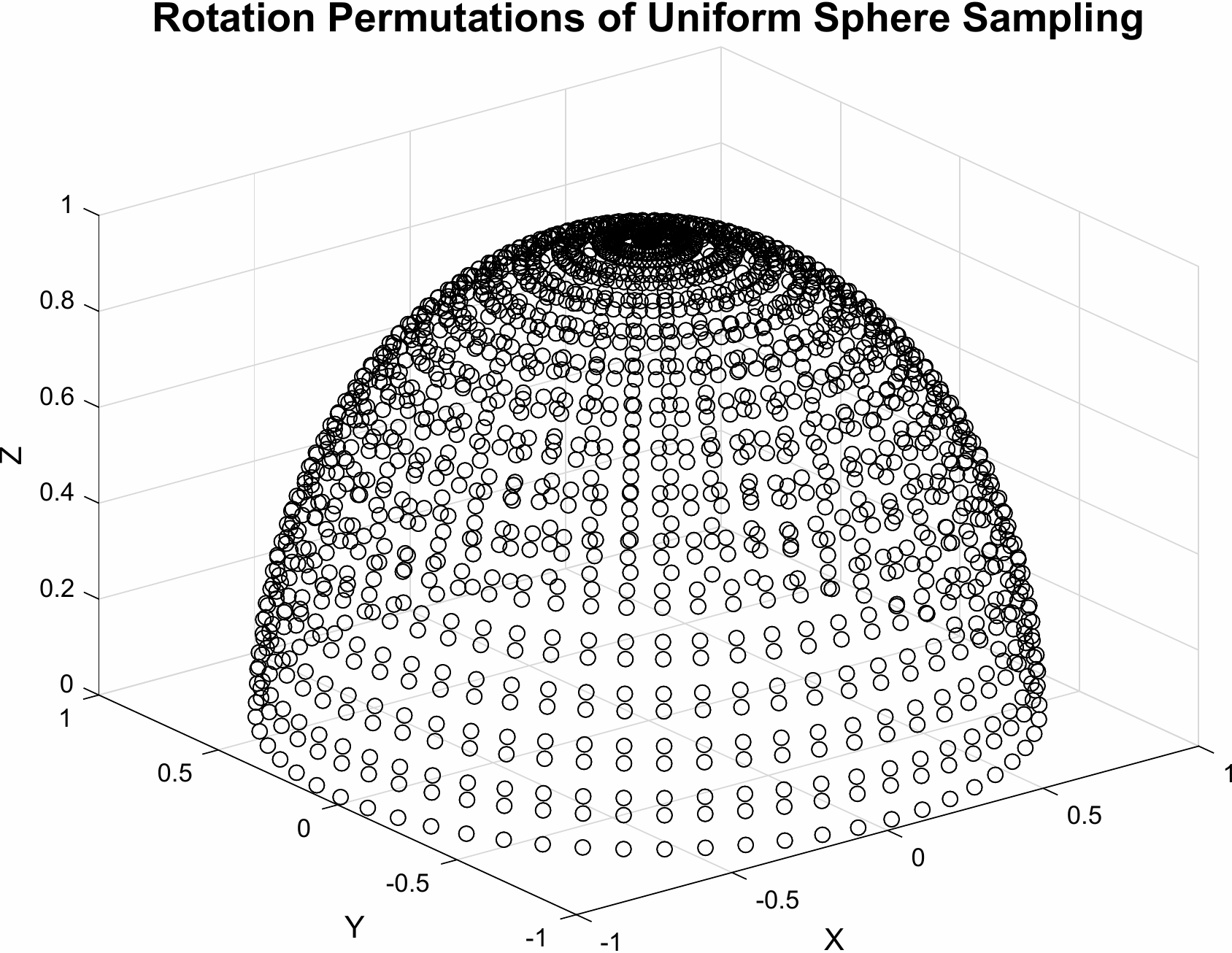}
		\caption{Uniform}
		\label{fig:rp1}
	\end{minipage}
	\begin{minipage}[b]{0.3\textwidth}
		\includegraphics[width=\textwidth]{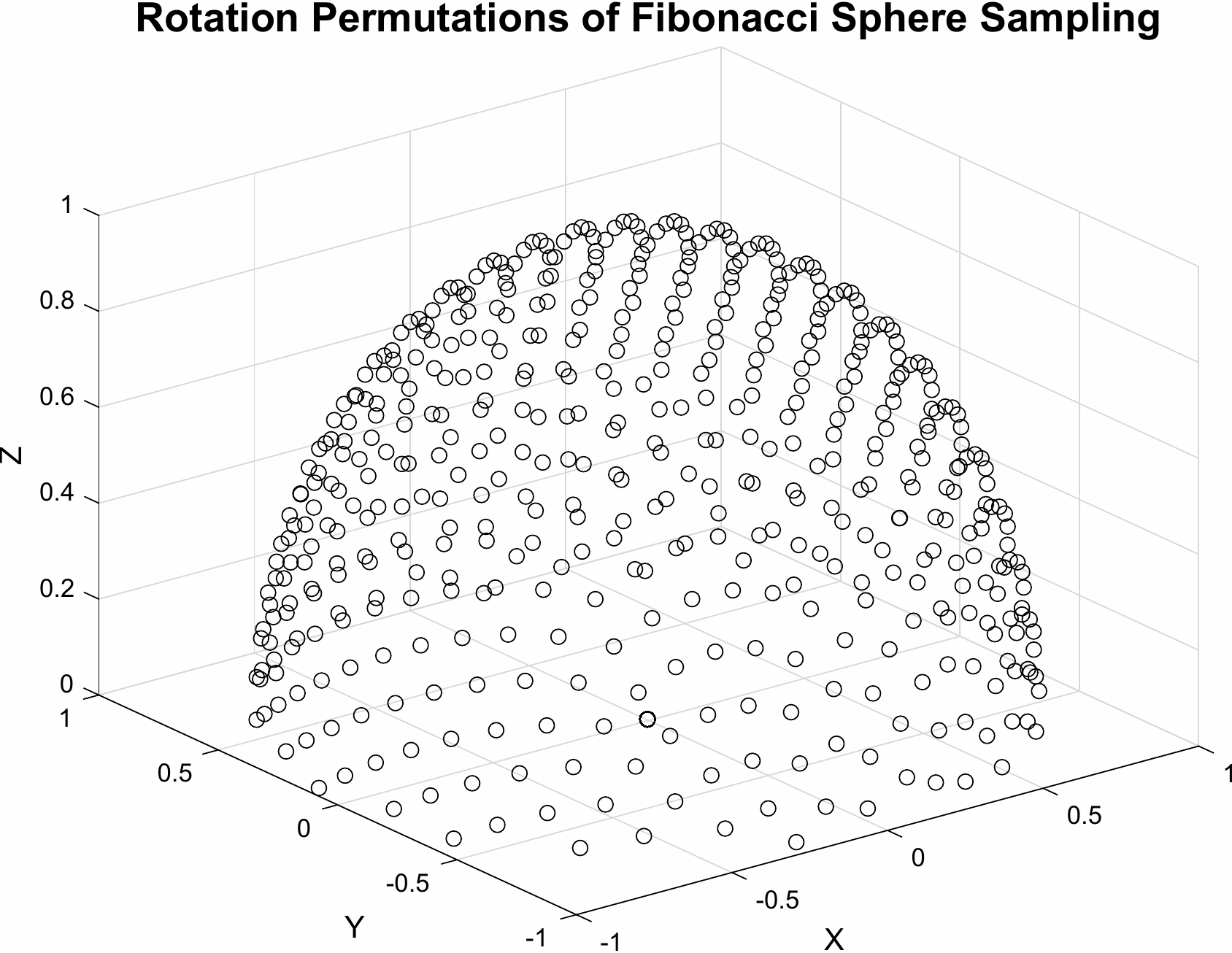}
		\caption{Fibonacci}
		\label{fig:rp2}
	\end{minipage}
	\begin{minipage}[b]{0.3\textwidth}
		\includegraphics[width=\textwidth]{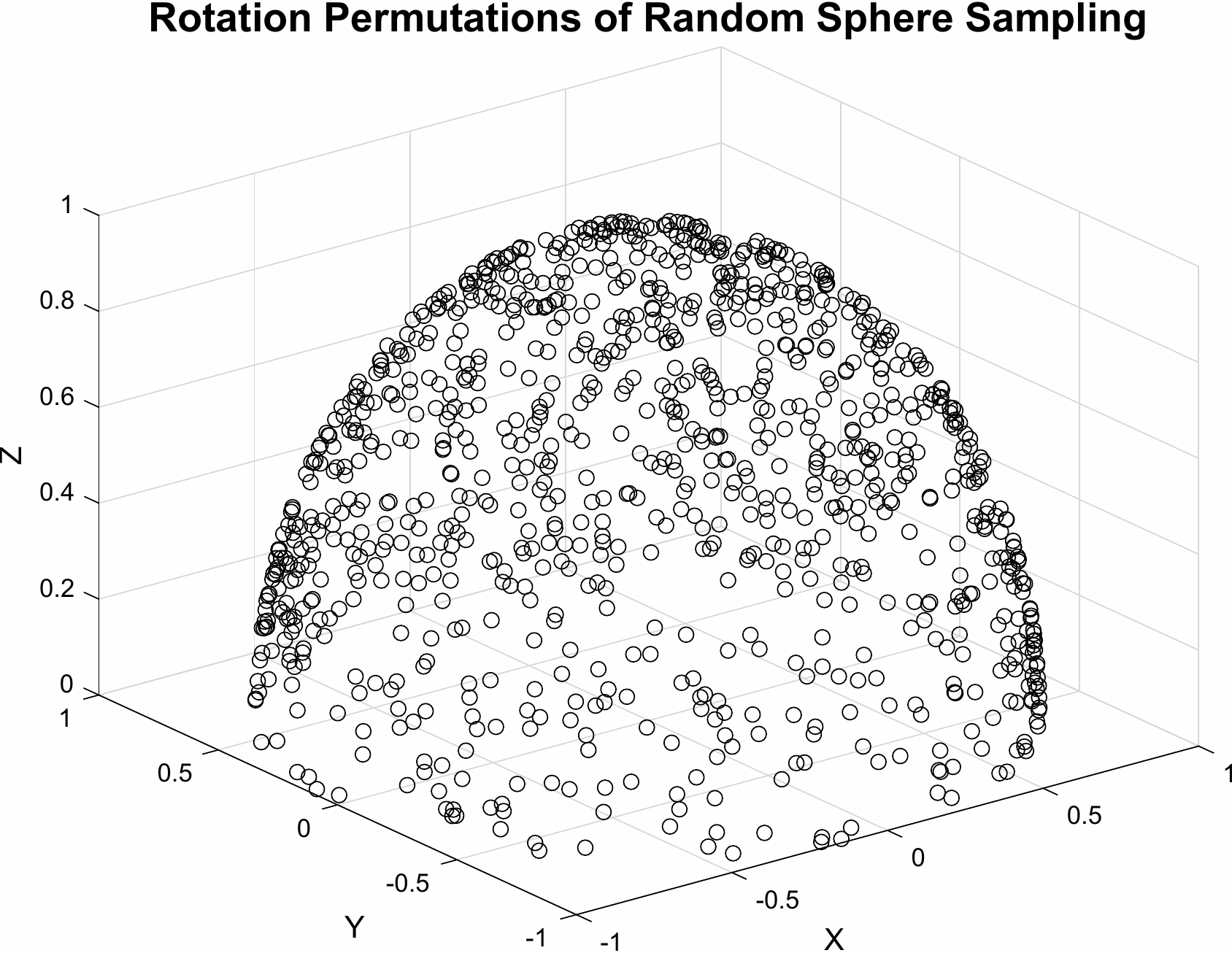}
		\caption{Random}
		\label{fig:rp3}
	\end{minipage}
\end{figure}

Figures \ref{fig:rp1} to \ref{fig:rp3} illustrate our data generation sampling strategies and shows the new Z-axis, \emph{i.e.} normals, of the sampling planes with respect to the origin. 

\subsection{Cartesian co-ordinates to rotation and offset} 

To calculate the rotation error of the predicted plane, we use linear algebra to create the rotation matrix. The predicted Cartesian points are likely to not form a perfect isosceles triangle formation compared to the ground truth. We exclusively use $p_c$ as the plane's origin point in world space with plane rotation calculated by the following method.

\begin{lstlisting}[language=Python]
def calculate_R(p1,p2,p3):
v1 = p3 - p1
v2 = p2 - p1
n1 = np.cross(v1,v2)
n2 = np.cross(n1,v1)
v1_norm = v1 / np.linalg.norm(v1) #x
n2_norm = n2 / np.linalg.norm(n2) #y
n1_norm = n1 / np.linalg.norm(n1) #z
R_recon = [[v1_norm[0] , n2_norm[0] , n1_norm[0]], 
[v1_norm[1] , n2_norm[1] , n1_norm[1]], 
[v1_norm[2] , n2_norm[2] , n1_norm[2]]]
return R_recon
\end{lstlisting}

\subsection{Euler and Quaternion Loss Functions} 

Any function, given that it's differentiable, can be used as a layer in a neural network. The distance metrics for 3D rotation are differentiable. Therefore, we are able to implement it as a network loss layer for regressing angle parameters.

\subsubsection{Euler Loss Function} 

\begin{equation}
\label{eq:phi1}
\begin{aligned}
{ \Phi  }_{ 1 }\left( \left( { \alpha  }_{ 1 },{ \beta  }_{ 1 },{ \gamma  }_{ 1 } \right) ,\left( { \alpha  }_{ 2 },{ \beta  }_{ 2 },{ \gamma  }_{ 2 } \right)  \right) &= \sqrt { { d\left( { \alpha  }_{ 1 },{ \alpha  }_{ 2 } \right)  }^{ 2 }+{ d\left( { \beta  }_{ 1 },{ \beta  }_{ 2 } \right)  }^{ 2 }+{ d\left( { \gamma  }_{ 1 },{ \gamma  }_{ 2 } \right)  }^{ 2 } } \\
\textrm{where} \quad d\left( a,b \right) &= \min { \left\{ \left| a-b \right| ,2\pi -\left| a-b \right|  \right\}  } 
\end{aligned}
\end{equation}

\subsubsection{Euler Back Propagation Function} 

\begin{equation}
\label{eq:phi1grad}
\begin{aligned}
\frac { \partial  }{ \partial { \alpha  }_{ 1 } } \left( { \Phi  }_{ 1 } \right) &= \frac { d\left( { \alpha  }_{ 1 },{ \alpha  }_{ 2 } \right)  }{ { \Phi  }_{ 1 } } \cdot \frac { \partial  }{ \partial { \alpha  }_{ 1 } } \left( d\left( { \alpha  }_{ 1 },{ \alpha  }_{ 2 } \right)  \right) \\
\frac { \partial  }{ \partial a } \left( d\left( a,b \right)  \right) &= { { -\textrm{sgn}\left( \left| a-b \right| -\pi  \right)  } }\cdot \textrm{sgn}\left( a-b \right) \\
\frac { \partial  }{ \partial b } \left( d\left( a,b \right)  \right) &= { {  \textrm{sgn}\left( \left| a-b \right| -\pi  \right)  } }\cdot \textrm{sgn}\left( a-b \right) 
\end{aligned}
\end{equation}

\subsubsection{Quaternion Loss Function} 

\begin{equation}
\label{eq:phi2}
{ \Phi  }_{ 2 }\left( { q }_{ 1 },{ q }_{ 2 } \right) =\min { \left\{ \left\| { q }_{ 1 }-{ q }_{ 2 } \right\| ,\left\| { q }_{ 1 }+{ q }_{ 2 } \right\|  \right\}  } 
\end{equation}

\subsubsection{Quaternion Back Propagation Function} 

\begin{equation}
\label{eq:phi2grad}
\begin{aligned}
\frac { \partial  }{ \partial { q }_{ 1 }^{ (1) } } \left( { \Phi  }_{ 2 }\left( { q }_{ 1 },{ q }_{ 2 } \right)  \right) &= \frac { 1 }{ 4 } \left( { q }_{ 1 }^{ (1) }+{ q }_{ 2 }^{ (1) }\textrm{sgn}\left( \left\{ \left\| { q }_{ 1 }-{ q }_{ 2 } \right\| ,\left\| { q }_{ 1 }+{ q }_{ 2 } \right\|  \right\} \right)  \right) \\
\frac { \partial  }{ \partial { q }_{ 2 }^{ (1) } } \left( { \Phi  }_{ 2 }\left( { q }_{ 1 },{ q }_{ 2 } \right)  \right) &= \frac { 1 }{ 4 } \left( { q }_{ 2 }^{ (1) }+{ q }_{ 1 }^{ (1) }\textrm{sgn}\left( \left\{ \left\| { q }_{ 1 }-{ q }_{ 2 } \right\| ,\left\| { q }_{ 1 }+{ q }_{ 2 } \right\|  \right\} \right)  \right) 
\end{aligned}
\end{equation}

where $q_1 = \{ q_1^{(1)} , q_1^{(2)} , q_1^{(3)} , q_1^{(4)} \}$ and $q_2 = \{ q_2^{(1)} , q_2^{(2)} , q_2^{(3)} , q_2^{(4)} \}$

\subsection{Network details} 

Table~\ref{tab:network-topo} lists the details of the SVRNet architecture as it is shown in Figure 1 in the paper in a complementary textual way. 

\begin{table}[htbp]
	\centering
	\begin{tabular}{c|c|c|c|c|c|c|cc}
		\toprule
		& \multicolumn{3}{c|}{ Layer Type ~} & ~ Kernel , Stride , Pad ~~      & \multicolumn{3}{c}{ ~ Parameters ~~}  \\
		\midrule
		Data Layer     & \multicolumn{3}{c|}{256x256x1}     & -                               & \multicolumn{3}{c}{-}                 \\
		\midrule
		\multirow{10}{*}{\begin{tabular}[c]{@{}c@{}}Convolutional \\Layers\end{tabular}}
		& \multicolumn{3}{c|}{conv1-96}      & 11,4,0                          & \multicolumn{3}{c}{11712}             \\ 
		& \multicolumn{3}{c|}{pool1}         & 3,2,0                           & \multicolumn{3}{c}{-}                 \\ 
		& \multicolumn{3}{c|}{LRN1}          & -                               & \multicolumn{3}{c}{-}                 \\ 
		& \multicolumn{3}{c|}{conv2-256}     & 5,1,2                           & \multicolumn{3}{c}{307456}            \\ 
		& \multicolumn{3}{c|}{pool2}         & 3,2,0                           & \multicolumn{3}{c}{-}                 \\ 
		& \multicolumn{3}{c|}{LRN2}          & -                               & \multicolumn{3}{c}{-}                 \\ 
		& \multicolumn{3}{c|}{conv3-384}     & 3,1,1                           & \multicolumn{3}{c}{885120}            \\ 
		& \multicolumn{3}{c|}{conv4-384}     & 3,1,1                           & \multicolumn{3}{c}{663936}            \\ 
		& \multicolumn{3}{c|}{conv5-256}     & 3,1,1                           & \multicolumn{3}{c}{442624}            \\ 
		& \multicolumn{3}{c|}{pool5}         & 3,2,0                           & \multicolumn{3}{c}{-}                 \\ 
		\midrule
		\multirow{3}{*}{\begin{tabular}[c]{@{}c@{}}Fully Connected \\Layers\end{tabular}} 
		& \multicolumn{3}{c|}{fc6}           & -                               & \multicolumn{3}{c}{51384320}          \\ 
		& \multicolumn{3}{c|}{fc7}           & -                               & \multicolumn{3}{c}{16781312}          \\ 
		& \multicolumn{3}{c|}{fc8}           & -                               & \multicolumn{3}{c}{4097000}           \\ 
		\midrule
		~~ Regression Layers ~~
		& ~ p1 ~ & ~ p2 ~ & ~ p3 ~           & -                               & 3003 & 3003 & 3003                    \\
		\bottomrule
	\end{tabular}
	\caption{SVRNet Network Topology }
	\label{tab:network-topo}
\end{table}

\subsection{Randomly selected illustrative inference results}

Here we show for Exp. 1, Exp. 2, and Exp. 3 randomly selected examples of images that have been presented to the network (ground truth) compared to an image sampled at the predicted location.

In these experiments, we present a ground truth (GT) image to the network to estimate the respective  transformation parameters needed to reorient the slice in its correct world co-ordinates. Using the transformation parameters, we generated a slice from the 3D atlas in the location where the network has predicted that slice should be (denoted as SVRNet).

The slices are compared side-by-side to give a visual representation of ``where the slice really is'' and ``where the network thinks the slice is''. 

\subsubsection{ Exp. 1:}

\begin{figure}[!h]
	\centering
	\vspace{-0.5cm}
	\subfloat[][GT]{\includegraphics[height=2cm]{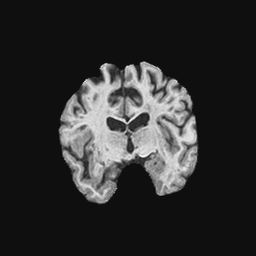}\label{}} 
	\subfloat[][SVRNet]{\includegraphics[height=2cm]{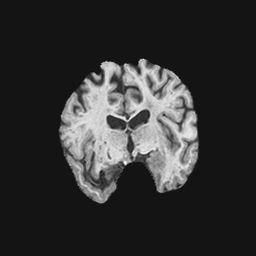}\label{}}
	\subfloat[][GT]{\includegraphics[height=2cm]{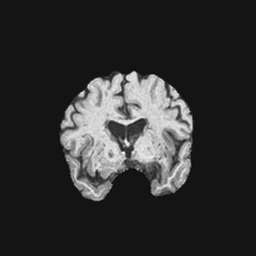}\label{}}
	\subfloat[][SVRNet]{\includegraphics[height=2cm]{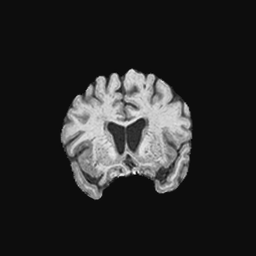}\label{}}
	\subfloat[][GT]{\includegraphics[height=2cm]{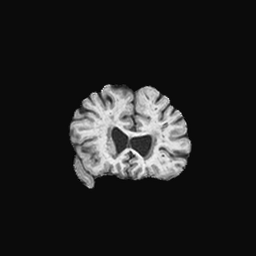}\label{}}
	\subfloat[][SVRNet]{\includegraphics[height=2cm]{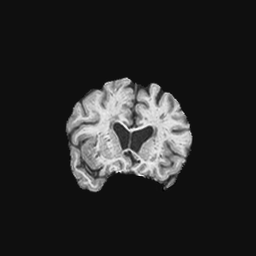}\label{}} \\
	
	\subfloat[][GT]{\includegraphics[height=2cm]{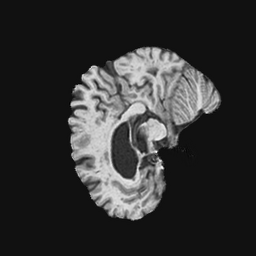}\label{}} 
	\subfloat[][SVRNet]{\includegraphics[height=2cm]{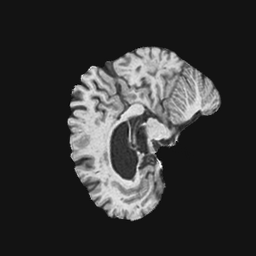}\label{}}	\subfloat[][GT]{\includegraphics[height=2cm]{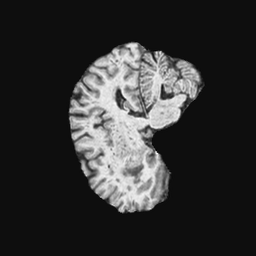}\label{}}
	\subfloat[][SVRNet]{\includegraphics[height=2cm]{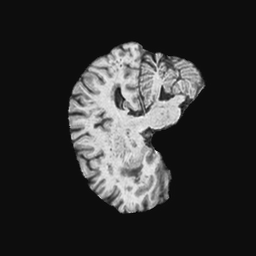}\label{}}
	\subfloat[][GT]{\includegraphics[height=2cm]{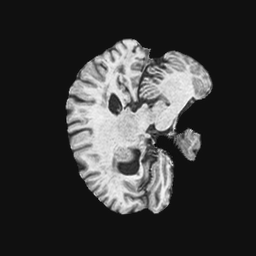}\label{}}
	\subfloat[][SVRNet]{\includegraphics[height=2cm]{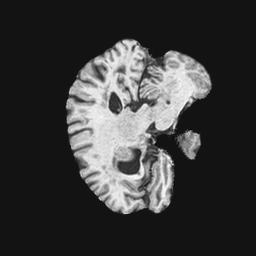}\label{}} \\
	
	\subfloat[][GT]{\includegraphics[height=2cm]{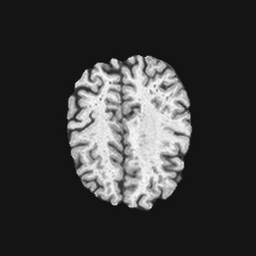}\label{}} 
	\subfloat[][SVRNet]{\includegraphics[height=2cm]{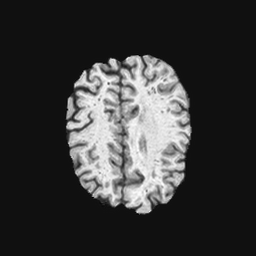}\label{}}
	\subfloat[][GT]{\includegraphics[height=2cm]{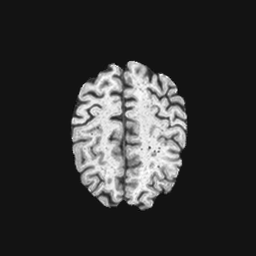}\label{}}
	\subfloat[][SVRNet]{\includegraphics[height=2cm]{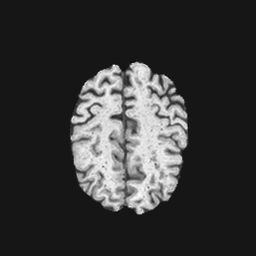}\label{}}
	\subfloat[][GT]{\includegraphics[height=2cm]{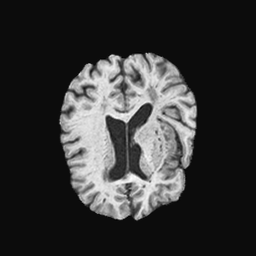}\label{}}
	\subfloat[][SVRNet]{\includegraphics[height=2cm]{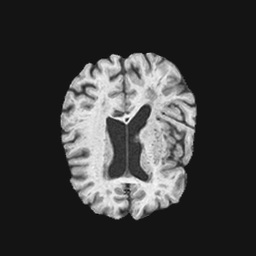}\label{}}
	
	\caption{ Exp 1: Examples of correct Ground Truth (GT) vs. Predicted (SVRNet) slice transformations. }
	\label{fig:reconADNIComp_Good}
\end{figure}

\begin{figure}[!h]
	\centering
	\vspace{-0.5cm}
	\subfloat[][GT]{\includegraphics[height=2cm]{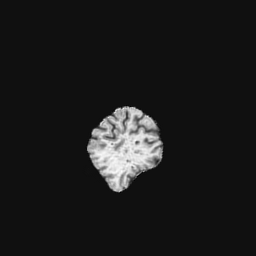}\label{}}
	\subfloat[][SVRNet1]{\includegraphics[height=2cm]{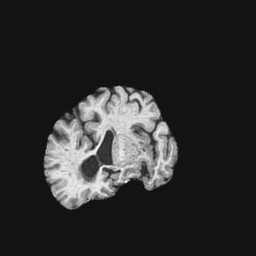}\label{}}
	\subfloat[][GT]{\includegraphics[height=2cm]{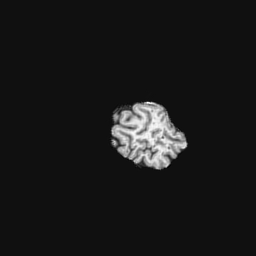}\label{}}
	\subfloat[][SVRNet]{\includegraphics[height=2cm]{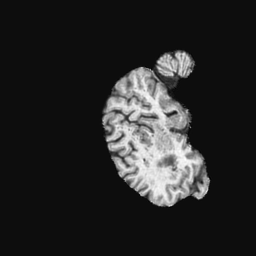}\label{}}
	\subfloat[][GT]{\includegraphics[height=2cm]{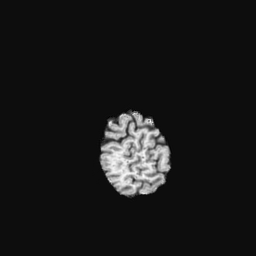}\label{}}
	\subfloat[][SVRNet]{\includegraphics[height=2cm]{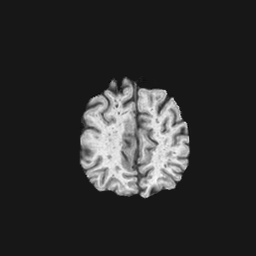}\label{}}
	\caption{ Exp 1: Examples of incorrect Ground Truth (GT) vs. Predicted (SVRNet) slice transformations. Note that mainly images at the boundaries of the trained region of interest have been incorrectly predicted.}
	\label{fig:reconADNIComp_Bad}
\end{figure}

\begin{figure}[!h]
	\centering
	\vspace{-0.5cm}
	\subfloat[][Test 1]{\includegraphics[height=3cm]{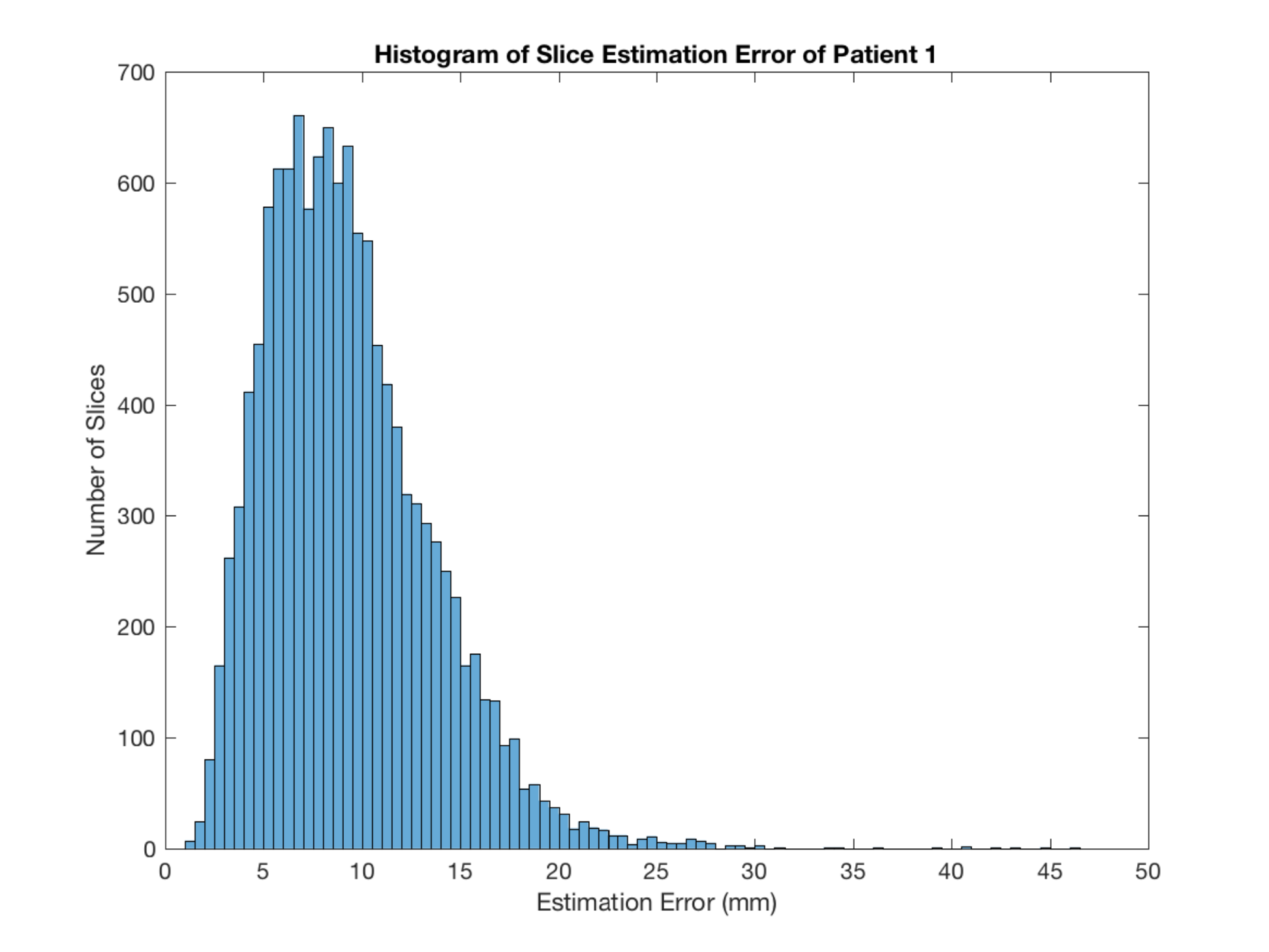}\label{}} \hfill
	\subfloat[][Test 2]{\includegraphics[height=3cm]{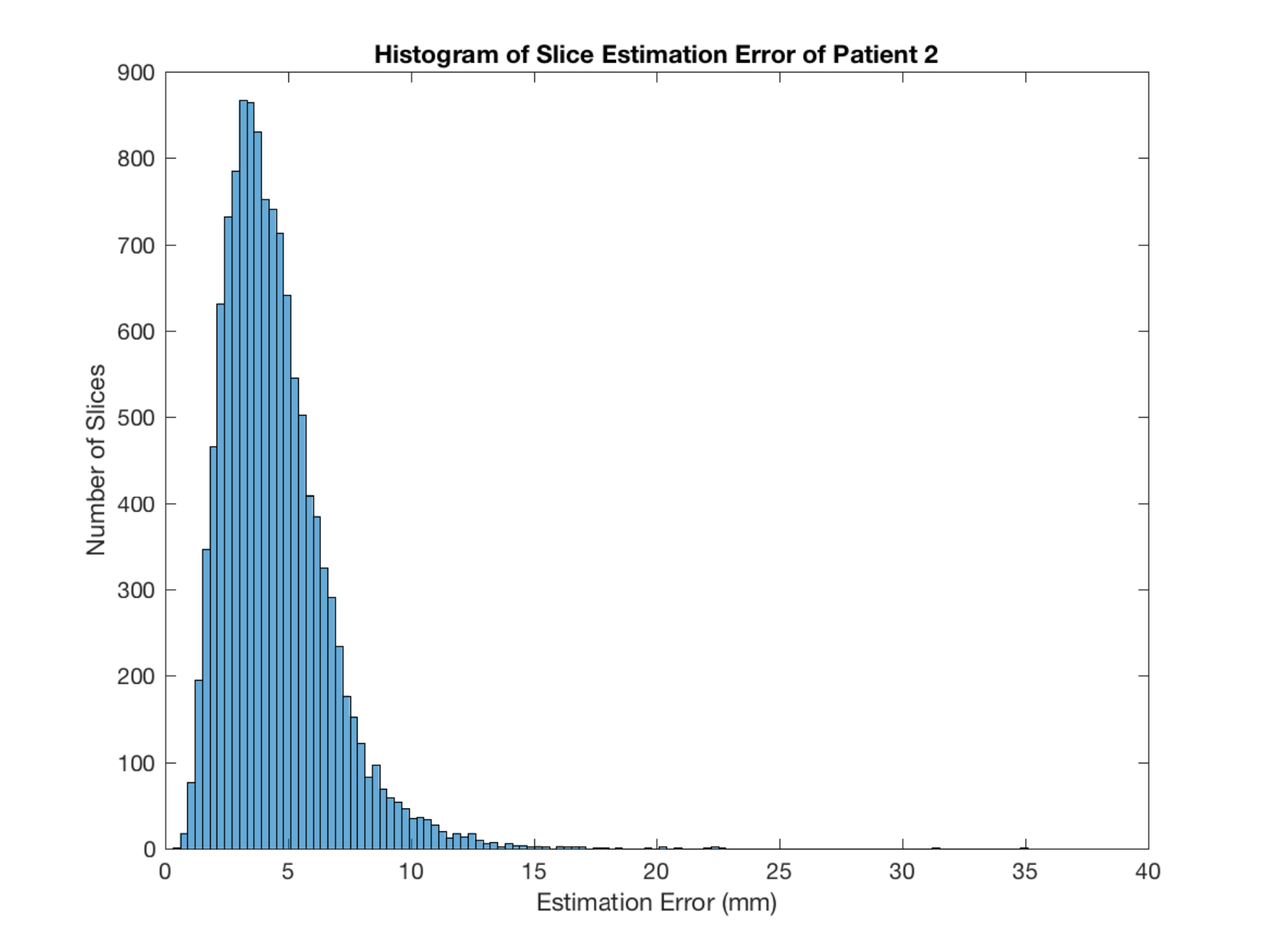}\label{}} \hfill
	\subfloat[][Test 3]{\includegraphics[height=3cm]{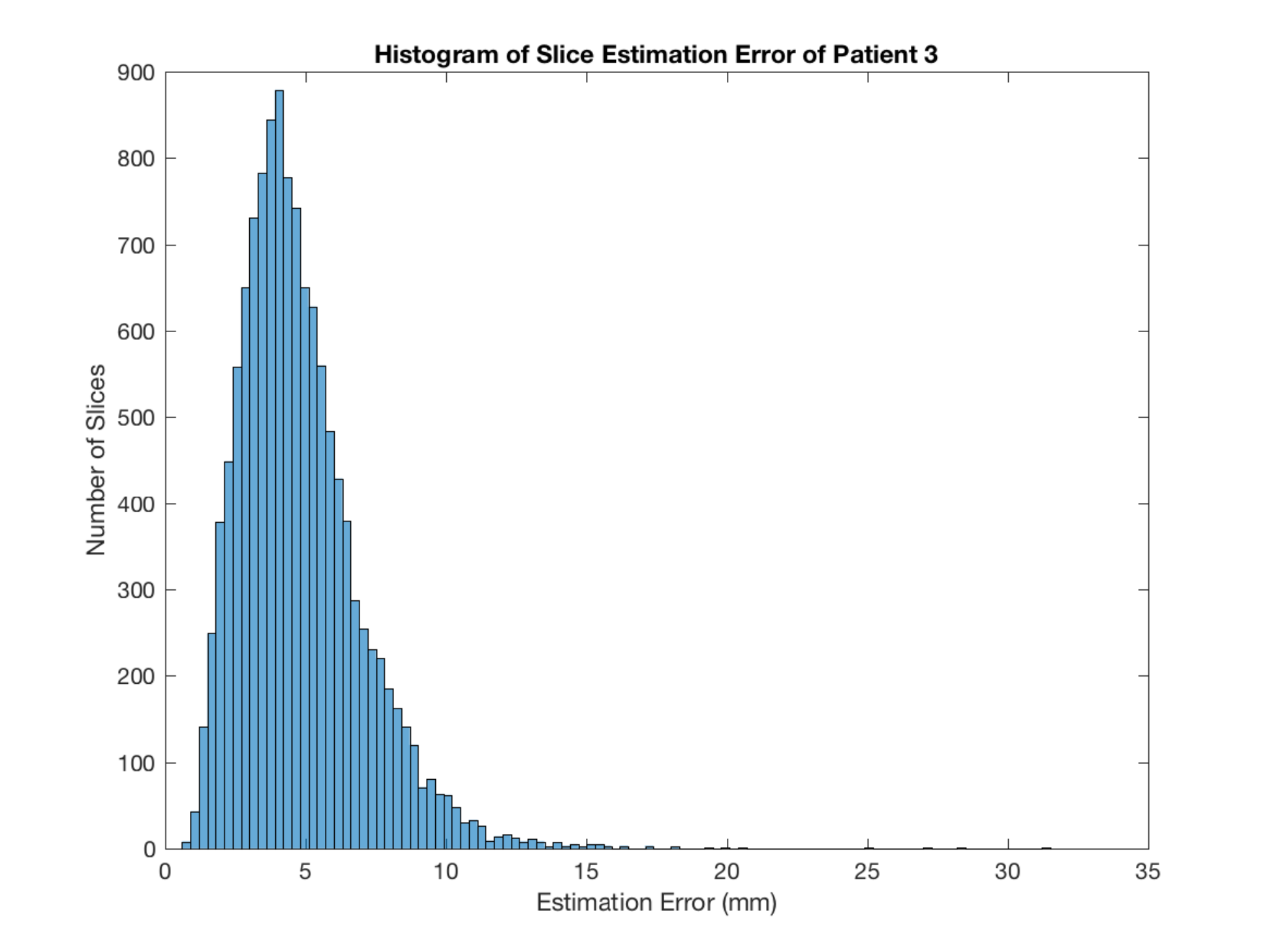}\label{}} \hfill
	\caption{ Exp 1: Histogram of three example volumes to show distribution of prediction error (in mm) with slices from the adult brain validation set. }
	\label{fig:adniHistograms}
	\vspace{-0.8cm}
\end{figure}

Slices, extracted from a correctly registered and reconstructed 3D volume, from the testing data set are presented to the network. The predicted slice is extracted from the same volume, using parameters estimated by SVRNet as shown in Fig.~\ref{fig:reconADNIComp_Good} and \ref{fig:reconADNIComp_Bad}.

\newpage
\subsubsection{ Exp. 2:}

\begin{figure}[!h]
	\centering
	\subfloat[][GT]{\includegraphics[height=2cm]{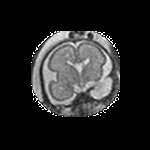}\label{}} 
	\subfloat[][SVRNet]{\includegraphics[height=2cm]{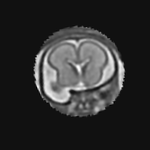}\label{}}
	\subfloat[][GT]{\includegraphics[height=2cm]{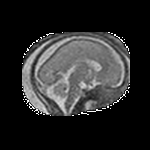}\label{}}
	\subfloat[][SVRNet]{\includegraphics[height=2cm]{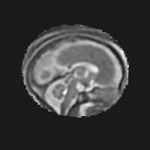}\label{}}
	\subfloat[][GT3]{\includegraphics[height=2cm]{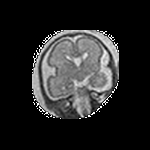}\label{}}
	\subfloat[][SVRNet]{\includegraphics[height=2cm]{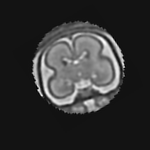}\label{}} \\
	
	\subfloat[][GT]{\includegraphics[height=2cm]{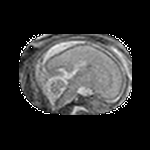}\label{}} 
	\subfloat[][SVRNet]{\includegraphics[height=2cm]{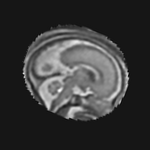}\label{}}
	\subfloat[][GT]{\includegraphics[height=2cm]{imgout_15_image.png}\label{}}
	\subfloat[][SVRNet]{\includegraphics[height=2cm]{imgout_15_estGT.png}\label{}}
	\subfloat[][GT]{\includegraphics[height=2cm]{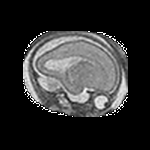}\label{}}
	\subfloat[][SVRNet]{\includegraphics[height=2cm]{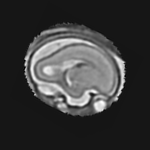}\label{}} \\
	
	\subfloat[][GT]{\includegraphics[height=2cm]{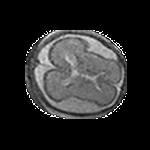}\label{}} 
	\subfloat[][SVRNet]{\includegraphics[height=2cm]{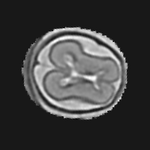}\label{}}
	\subfloat[][GT]{\includegraphics[height=2cm]{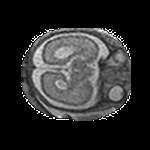}\label{}}
	\subfloat[][SVRNet]{\includegraphics[height=2cm]{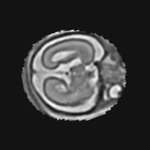}\label{}}
	\subfloat[][GT]{\includegraphics[height=2cm]{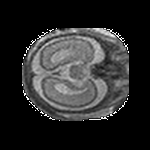}\label{}}
	\subfloat[][SVRNet]{\includegraphics[height=2cm]{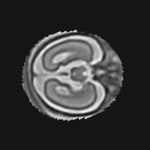}\label{}} \\
	
	\caption{ Exp 2: Examples of correct Ground Truth (GT) vs. Predicted (SVRNet) slice transformations. }
	\label{fig:reconFetalComp_Good}
\end{figure}

\begin{figure}[!h]
	\centering
	\subfloat[][GT]{\includegraphics[height=2cm]{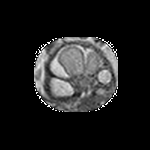}\label{}}
	\subfloat[][SVRNet]{\includegraphics[height=2cm]{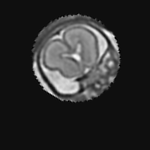}\label{}}
	\subfloat[][GT]{\includegraphics[height=2cm]{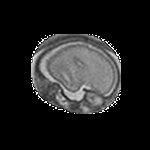}\label{}}
	\subfloat[][SVRNet]{\includegraphics[height=2cm]{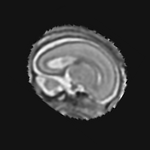}\label{}}
	\subfloat[][GT]{\includegraphics[height=2cm]{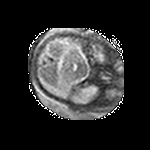}\label{}}
	\subfloat[][SVRNet]{\includegraphics[height=2cm]{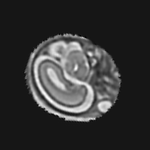}\label{}}
	\caption{ Exp 2: Examples of incorrect Ground Truth (GT) vs. Predicted (SVRNet) slice transformations. }
	\label{fig:reconFetalComp_Bad}
\end{figure}

Slices, from a motion corrupted MRI stack, are segmented and cropped. Since there is no ground truth for the queried images, an arbitrary fetal atlas is used for visualization in Fig.~\ref{fig:reconFetalComp_Good} and \ref{fig:reconFetalComp_Bad}.

\newpage
\subsubsection{ Exp. 3:}

\begin{figure}[!h]
	\centering
	\vspace{-0.5cm}
	\subfloat[][GT]{\includegraphics[height=2cm]{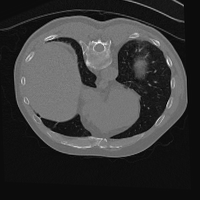}\label{}} 
	\subfloat[][SVRNet]{\includegraphics[height=2cm]{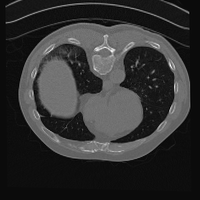}\label{}}
	\subfloat[][GT]{\includegraphics[height=2cm]{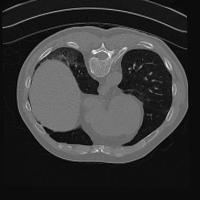}\label{}}
	\subfloat[][SVRNet]{\includegraphics[height=2cm]{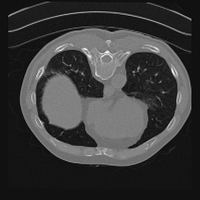}\label{}}
	\subfloat[][GT]{\includegraphics[height=2cm]{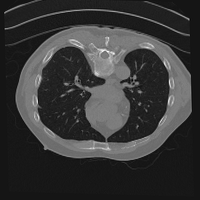}\label{}}
	\subfloat[][SVRNet]{\includegraphics[height=2cm]{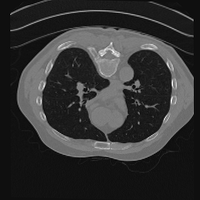}\label{}} \\
	
	\subfloat[][GT]{\includegraphics[height=2cm]{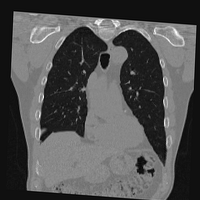}\label{}} 
	\subfloat[][SVRNet]{\includegraphics[height=2cm]{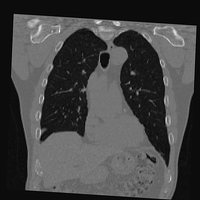}\label{}}
	\subfloat[][GT]{\includegraphics[height=2cm]{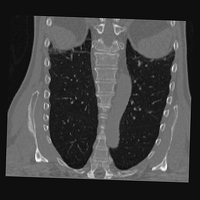}\label{}}
	\subfloat[][SVRNet]{\includegraphics[height=2cm]{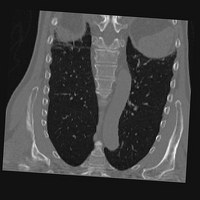}\label{}}
	\subfloat[][GT]{\includegraphics[height=2cm]{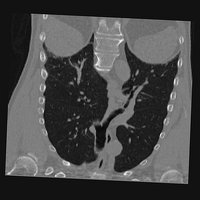}\label{}}
	\subfloat[][SVRNet]{\includegraphics[height=2cm]{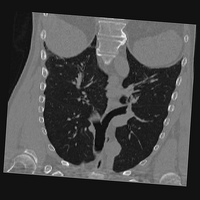}\label{}} \\
	
	\caption{ Exp 3: Examples of correct Ground Truth (GT) vs. Predicted (SVRNet) slice transformations.}
	\label{fig:reconThoraxComp_Good}
	\vspace{-0.75cm}
\end{figure}

\begin{figure}[!h]
	\centering
	\vspace{-0.75cm}
	\subfloat[][GT]{\includegraphics[height=2cm]{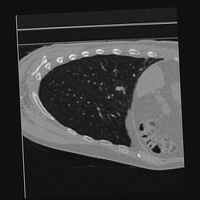}\label{}}
	\subfloat[][SVRNet]{\includegraphics[height=2cm]{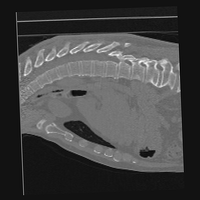}\label{}}
	\subfloat[][GT]{\includegraphics[height=2cm]{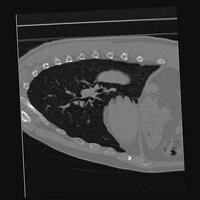}\label{}}
	\subfloat[][SVRNet]{\includegraphics[height=2cm]{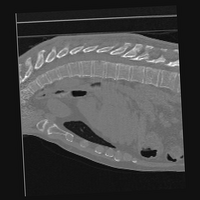}\label{}}
	\subfloat[][GT]{\includegraphics[height=2cm]{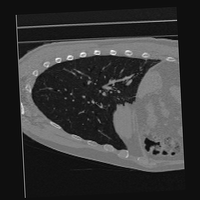}\label{}}
	\subfloat[][SVRNet]{\includegraphics[height=2cm]{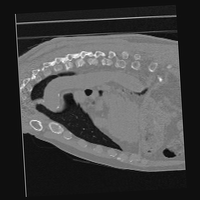}\label{}}
	\caption{ Exp 3: Examples of incorrect Ground Truth (GT) vs. Predicted (SVRNet) slice transformations. Note that most incorrect slices exist in the sagittal plane.}
	\label{fig:reconThoraxComp_Bad}
	\vspace{-0.75cm}
\end{figure}

\begin{figure}[!h]
	\centering
	\vspace{-0.75cm}
	\subfloat[][Test 1]{\includegraphics[height=3cm]{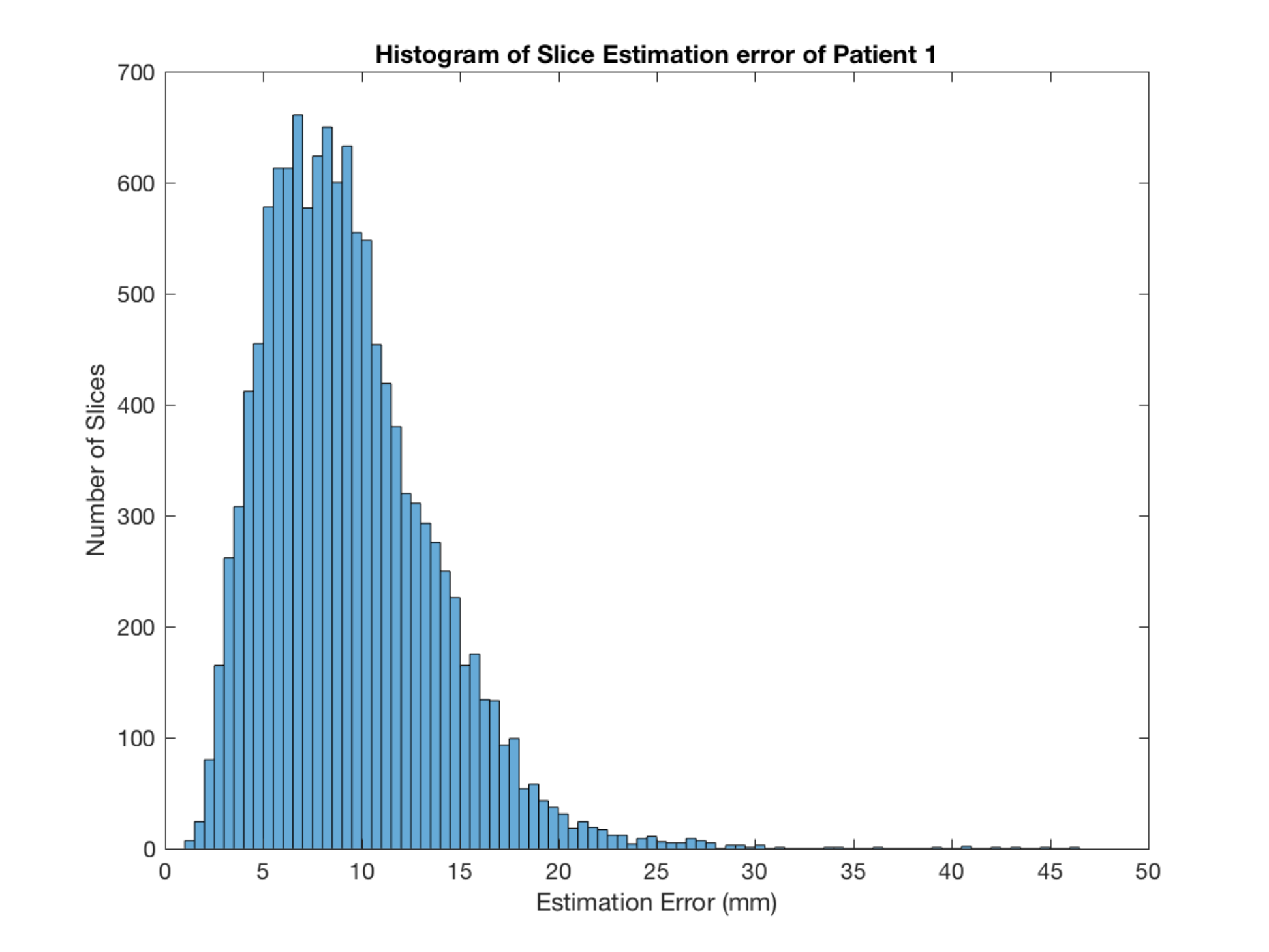}\label{}} \hfill
	\subfloat[][Test 2]{\includegraphics[height=3cm]{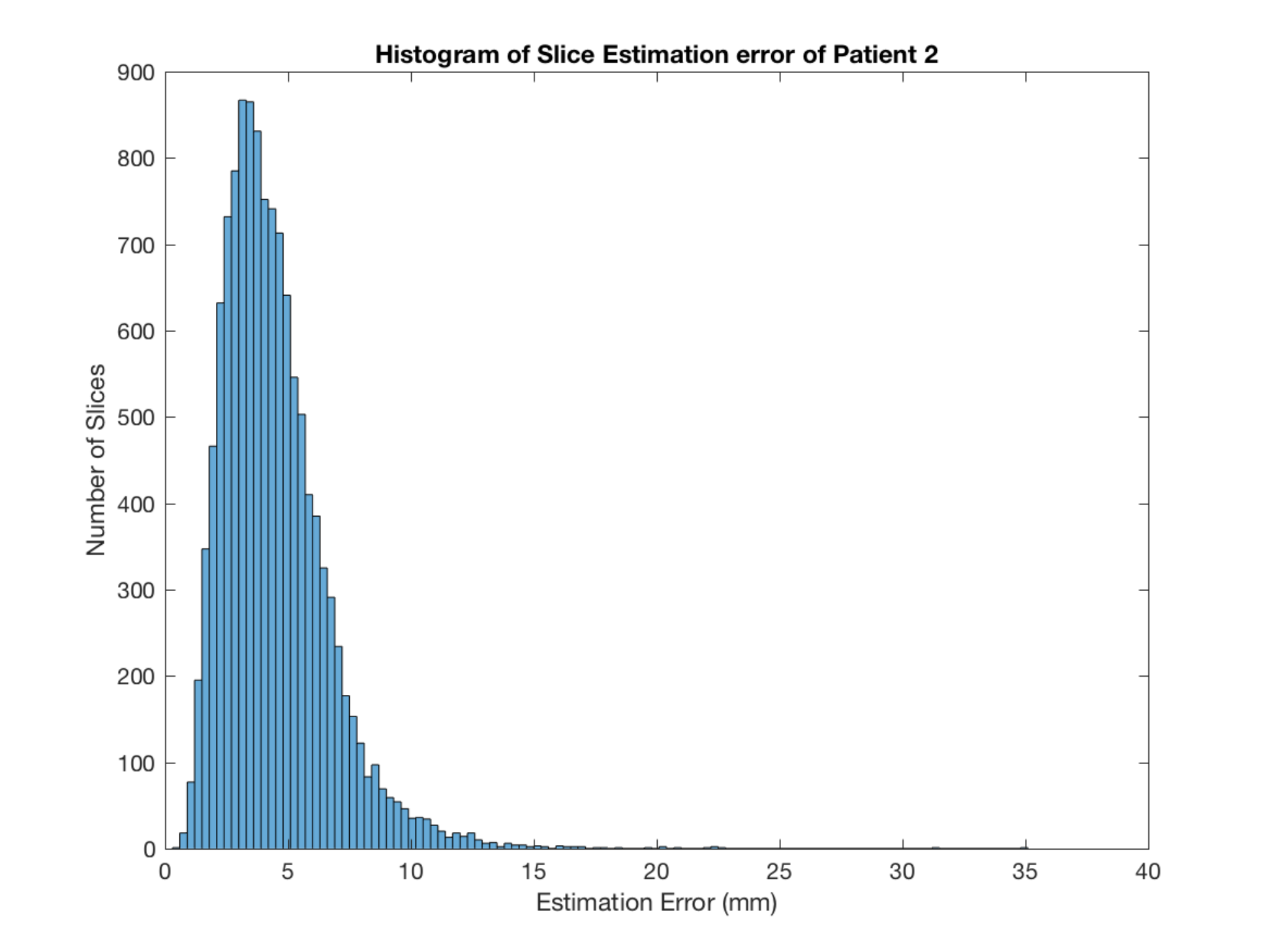}\label{}} \hfill
	\subfloat[][Test 3]{\includegraphics[height=3cm]{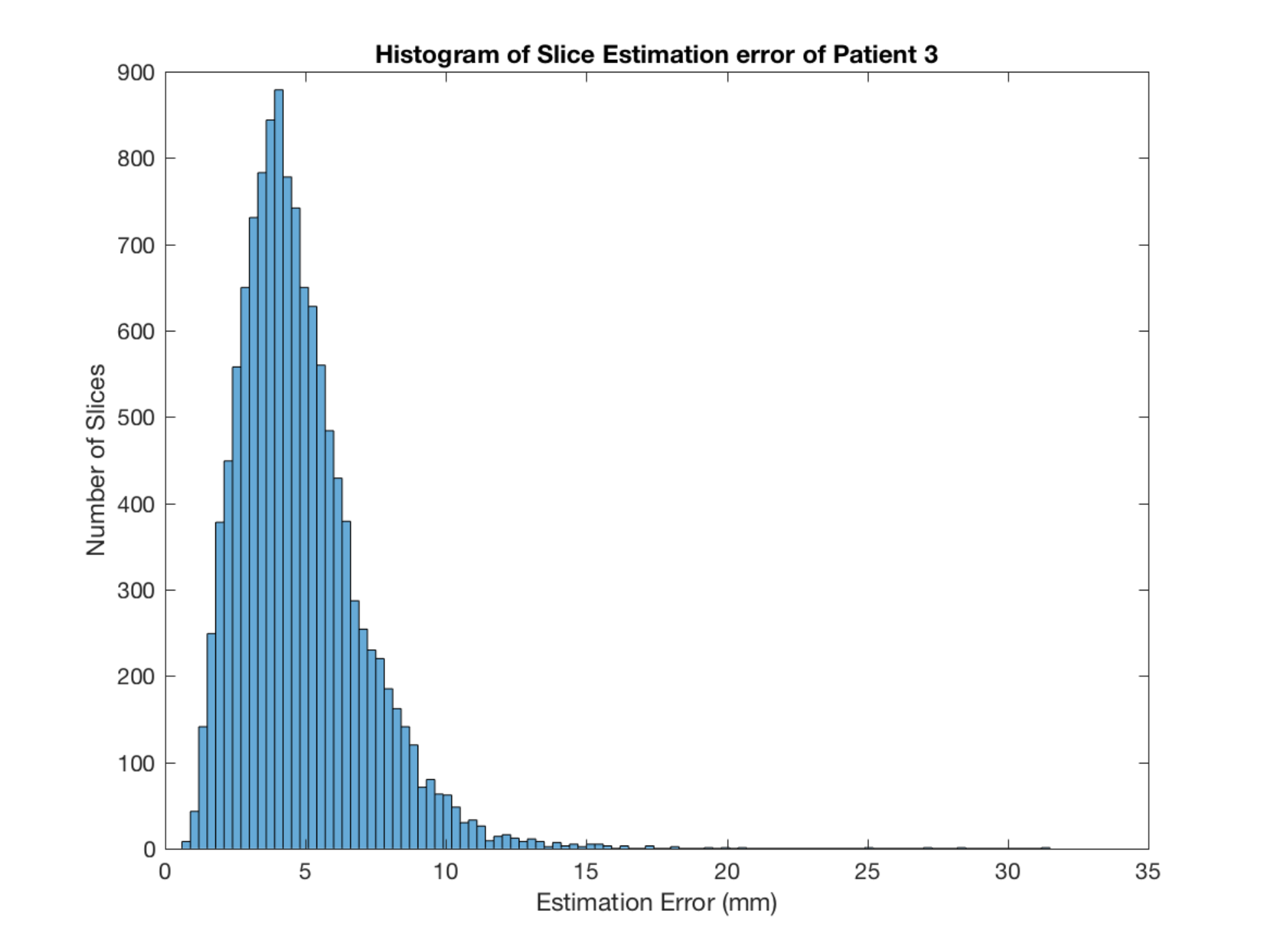}\label{}} \hfill
	\caption{ Exp 3: Histogram of three example volumes to show distribution of prediction error (in mm) with slices from the adult thorax validation set. }
	\label{fig:ThoraxHistograms}
	\vspace{-1cm}
\end{figure}

\begin{figure}[!h]
	\centering
	\subfloat[][GT]{\includegraphics[height=2cm]{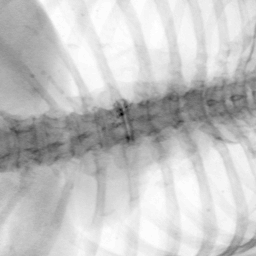}\label{}}
	\subfloat[][SVRNet]{\includegraphics[height=2cm]{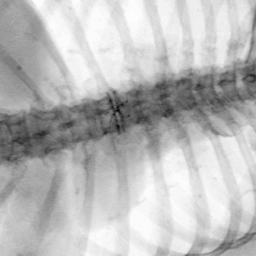}\label{}} \hfill
	\subfloat[][GT]{\includegraphics[height=2cm]{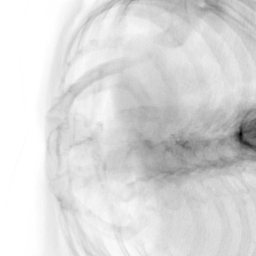}\label{}}
	\subfloat[][SVRNet]{\includegraphics[height=2cm]{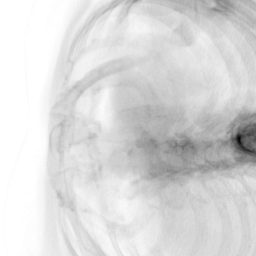}\label{}} \hfill
	\subfloat[][GT]{\includegraphics[height=2cm]{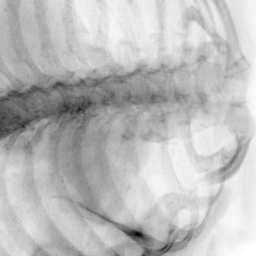}\label{}}
	\subfloat[][SVRNet]{\includegraphics[height=2cm]{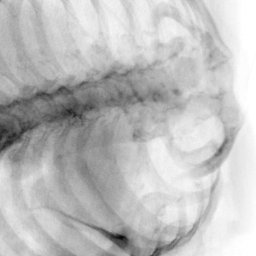}\label{}} \\
	
	\subfloat[][GT]{\includegraphics[height=2cm]{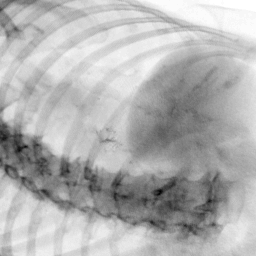}\label{}}
	\subfloat[][SVRNet]{\includegraphics[height=2cm]{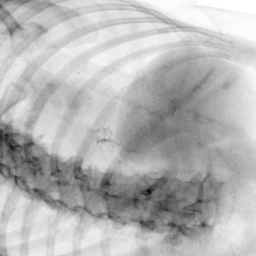}\label{}} \hfill
	\subfloat[][GT]{\includegraphics[height=2cm]{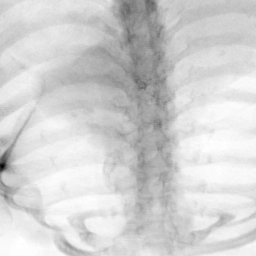}\label{}}
	\subfloat[][SVRNet]{\includegraphics[height=2cm]{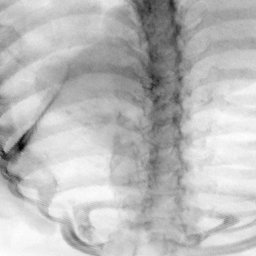}\label{}} \hfill
	\subfloat[][GT]{\includegraphics[height=2cm]{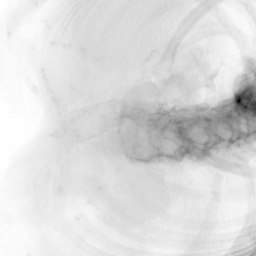}\label{}}
	\subfloat[][SVRNet]{\includegraphics[height=2cm]{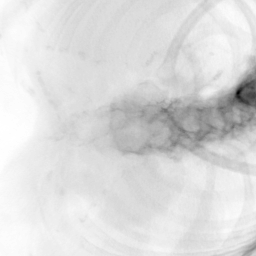}\label{}} \\
	
	
	\caption{ Exp 3: Randomly selected Ground Truth (GT) vs. Predicted (SVRNet) DRR locations.}
	\label{fig:DRRComp}
\end{figure}

\begin{figure}[!h]
	\centering
	\subfloat[][1000 healthy DRRs]{\includegraphics[height=4cm]{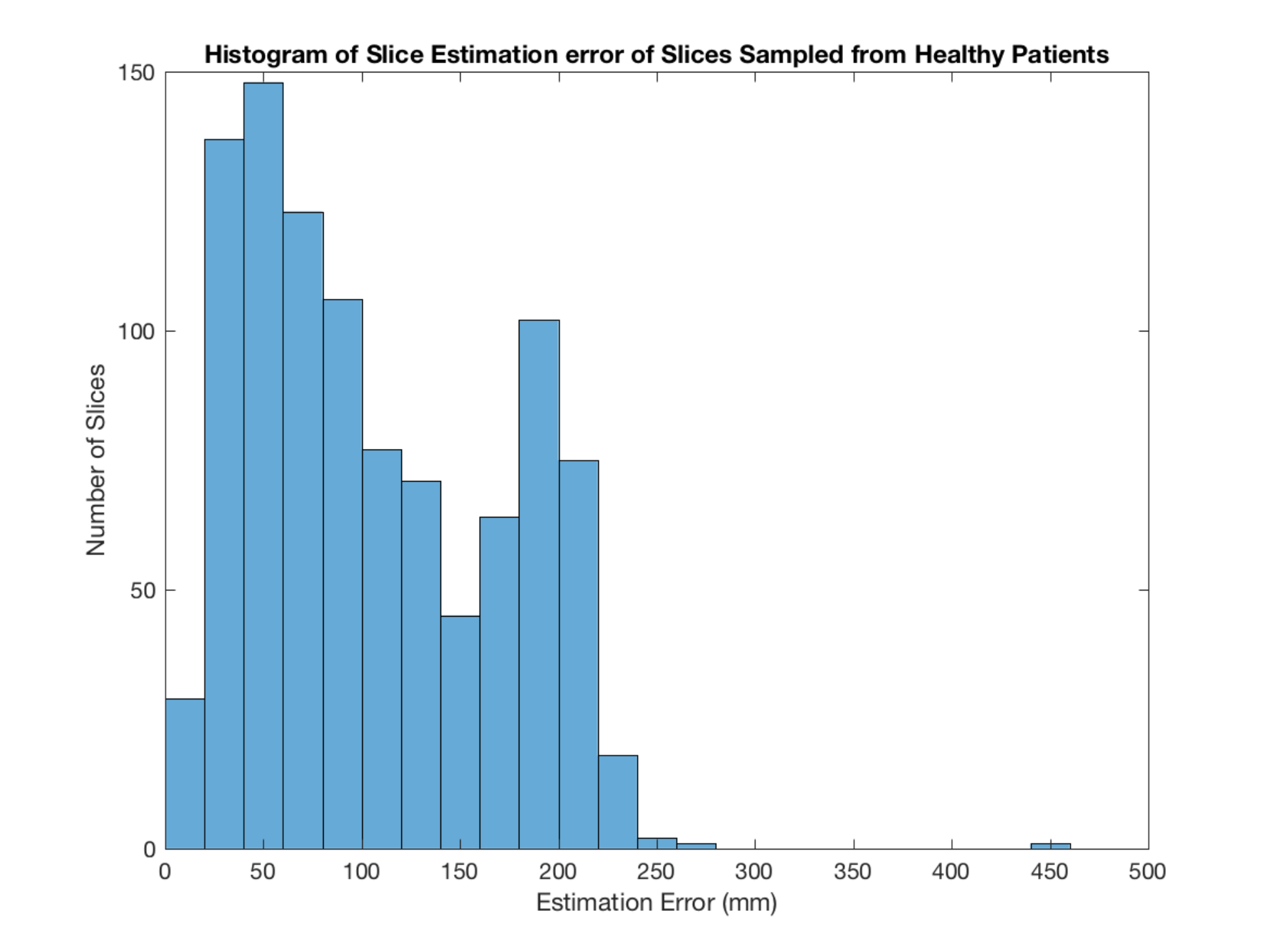}\label{}} \hfill
	\subfloat[][1000 pathological DRRs]{\includegraphics[height=4cm]{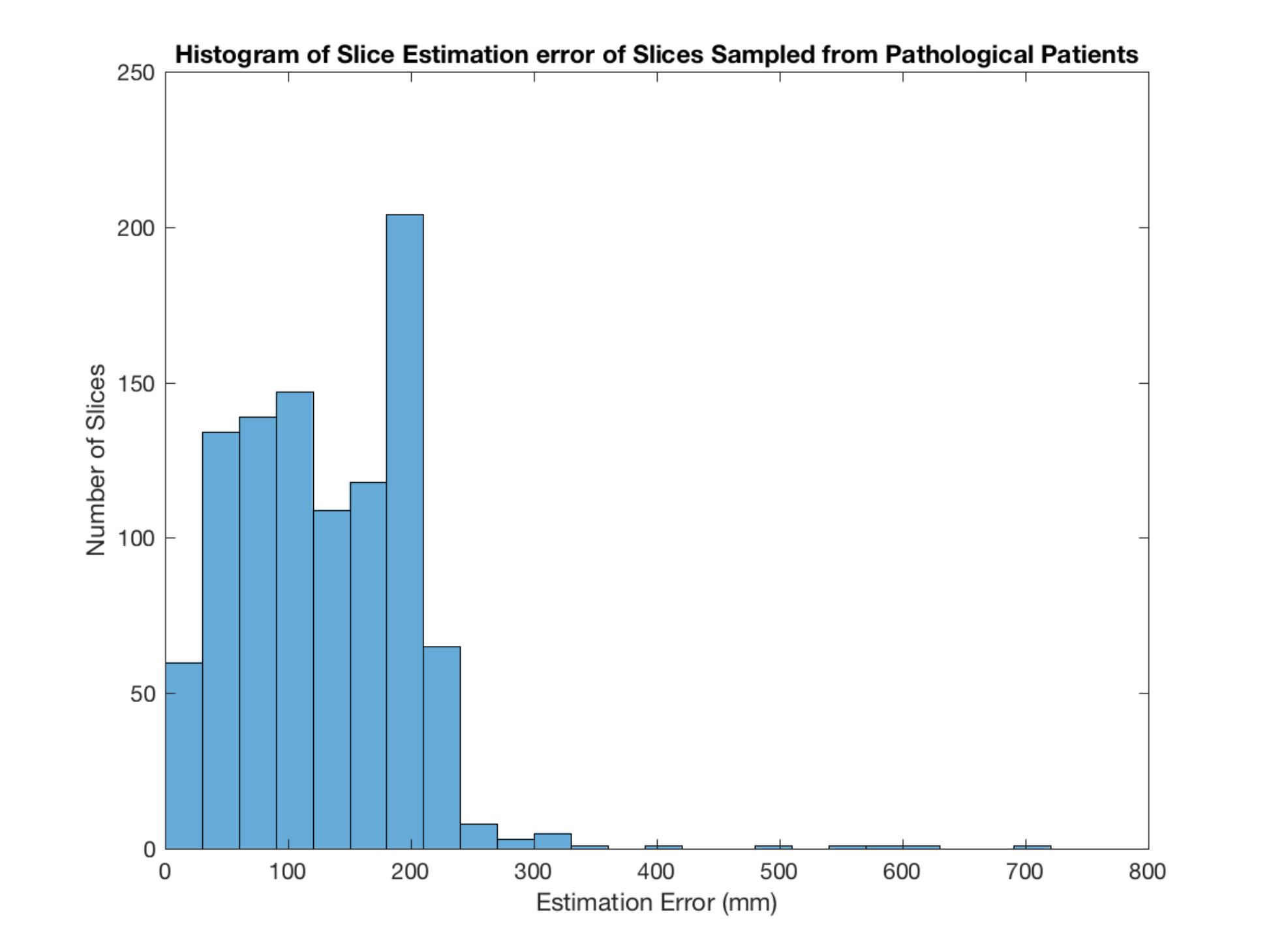}\label{}} \hfill
	\caption{ Exp 3: Histogram to show distribution of prediction error (in mm) with randomly selected Ground Truth DRR locations. }
	\label{fig:DRRHistograms}
\end{figure}

We replicated the experiment on adult thorax data without specifically segmented organs. This approach was applied to CT acquisition, shown in Fig~\ref{fig:reconThoraxComp_Good} and \ref{fig:reconThoraxComp_Bad}, as well as Digitally Reconstructed Radiographs generated using Siddon-Jacobs Ray Tracing shown in Fig.~\ref{fig:DRRComp}.

\end{document}